\documentclass[nohyperref]{article}
\PassOptionsToPackage{numbers, compress}{natbib}
\usepackage{iclr2025_conference, times}
\iclrfinalcopy

\usepackage[utf8]{inputenc} %

\usepackage{graphicx}

\usepackage[T1]{fontenc}

\usepackage[hidelinks,colorlinks]{hyperref}       %
\usepackage{url}            %
\usepackage{booktabs}       %
\usepackage{amsfonts}       %
\usepackage{nicefrac}       %
\usepackage{microtype}      %

\IfFileExists{headers/config/showoverfull.config}{
	\overfullrule=1cm
}{
}

\usepackage{marginnote}

\usepackage[backgroundcolor=none,linecolor=red,textsize=footnotesize]{todonotes}

\usepackage{etoolbox}

\newbool{includeappendix}
\setbool{includeappendix}{true} %
\IfFileExists{headers/config/noappendix.config}{
	\setbool{includeappendix}{false}
}{}

\newif\ifincludeappendixx
\ifbool{includeappendix}{
	\includeappendixxtrue
}{
	\includeappendixxfalse
}

\usepackage{xr} %
\usepackage{filecontents}

\ifbool{includeappendix}{}{
	\input{appendix-labels-loader}

	\externaldocument{appendix-labels}
}

\newcommand{\eg}{e.g., }
\newcommand{\ie}{i.e., }

\usepackage{acro} %

\usepackage{listings}

\usepackage{textcomp}

\usepackage{xcolor}

\usepackage[scaled=0.8]{beramono}

\definecolor{ckeyword}{HTML}{7F0055}
\definecolor{ccomment}{HTML}{3F7F5F}
\definecolor{cstring}{HTML}{2A0099}

\lstdefinestyle{numbers}{
	numbers=left,
	framexleftmargin=20pt,
	numberstyle=\tiny,
	firstnumber=auto,
	numbersep=1em,
	xleftmargin=2em
}

\lstdefinestyle{layout}{
	frame=none,
	captionpos=b,
}

\lstdefinestyle{comment-style}{
	morecomment=[l]//,
	morecomment=[s]{/*}{*/},
	commentstyle={\color{ccomment}\itshape},
}

\lstdefinestyle{string-style}{
	morestring=[b]",%
	morestring=[b]',%
	stringstyle={\color{cstring}},
	showstringspaces=false,%
}

\lstdefinestyle{keyword-style}{
	keywordstyle={\ttfamily\bfseries},
	morekeywords={
		function,
		constructor,
		int,
		bool,
		return,
		returns,
		uint
	},
	morekeywords = [2]{},
	keywordstyle = [2]{\text},
	sensitive=true,
}

\lstdefinestyle{input-encoding}{
	inputencoding=utf8,
	extendedchars=true,
	literate=
	{ℝ}{$\reals$}1%
	{→}{$\rightarrow$}1%
	{α}{$\alpha$}1%
	{β}{$\beta$}1%
	{λ}{$\lambda$}1%
	{θ}{$\theta$}1%
	{ϕ}{$\phi$}1%
}

\lstdefinestyle{escaping}{
	moredelim={**[is][\color{blue}]{\%}{\%}},
	escapechar=|,
	mathescape=true
}

\lstdefinestyle{default-style}{
	basicstyle=\fontencoding{T1}\ttfamily\footnotesize,
	style=numbers,
	style=layout,
	style=comment-style,
	style=string-style,
	style=keyword-style,
	style=input-encoding,
	style=escaping,
	tabsize=2,
	upquote=true
}

\lstdefinelanguage{BASIC}{
	language=C++,
	style=default-style
}[keywords,comments,strings]%

\usepackage[utf8]{inputenc} %
\usepackage[T1]{fontenc}    %
\usepackage{algcompatible}
\usepackage{caption}
\usepackage[most]{tcolorbox}
\usepackage{algpseudocode}
\usepackage{enumerate}
\usepackage{url}            %
\usepackage{booktabs}       %
\usepackage{amsfonts}       %
\usepackage{nicefrac}       %
\usepackage{microtype}      %
\usepackage{xcolor}         %
\usepackage[ruled,vlined]{algorithm2e}
\usepackage{tikz,booktabs,multirow,enumitem,bm}
\usepackage{colortbl}
\usepackage{tabularx}
\usepackage{subcaption}
\usepackage{wrapfig}
\usepackage{tabulary}
\usepackage{amsmath,amsthm,amsfonts, bbm}

\usepackage{amsmath,amsfonts,bm}
\usepackage{amsthm}

\def\1{\bm{1}}

\DeclareMathAlphabet{\mathsfit}{\encodingdefault}{\sfdefault}{m}{sl}
\SetMathAlphabet{\mathsfit}{bold}{\encodingdefault}{\sfdefault}{bx}{n}

\newcommand{\newdata}[0]{\textsc{Farad}}

\newcommand{\sib}[0]{\textsc{Sib}}
\newcommand{\ibm}[0]{\textsc{Aag}}
\newcommand{\facts}[0]{\textsc{Facts}}
\newcommand{\ours}[0]{\textsc{Ward}}
\newcommand{\problem}[0]{RAG-DI} 
\newcommand{\llm}[0]{\ensuremath{\mathcal{M}}}
\newcommand{\ragsys}[0]{\ensuremath{\mathcal{M}^\star}}
\newcommand{\ownerdata}[0]{\ensuremath{D_{\text{do}}}}

\newcommand{\repliqa}[0]{\emph{RepLiQA}}
\newcommand{\newdatae}[0]{\textsc{Farad}-\emph{Easy}}
\newcommand{\newdatah}[0]{\textsc{Farad}-\emph{Hard}} 

\newcommand{\gptBig}[0]{\textsc{GPT4o}}
\newcommand{\sonnet}[0]{\textsc{Claude3.5-Sonnet}}
\newcommand{\llamaBig}[0]{\textsc{Llama3.1-405B}}
\newcommand{\qwen}[0]{\textsc{Qwen1.5-110B}}

\newcommand{\gptSmall}[0]{\textsc{GPT3.5}}
\newcommand{\haiku}[0]{\textsc{Claude3-Haiku}}
\newcommand{\llamaMid}[0]{\textsc{Llama3.1-70B}}

\newcommand{\sentenceTf}[0]{\textsc{all-MiniLM-L6-v2}}
\newcommand{\llamaSmall}[0]{\textsc{Llama3.1-8B}}

\newcommand{\easy}[0]{\emph{Easy}}
\newcommand{\hard}[0]{\emph{Hard}}
\newcommand{\naive}[0]{\emph{Naive-P}}
\newcommand{\defended}[0]{\emph{Def-P}}

\newcommand{\incase}[0]{\emph{IN}}
\newcommand{\outcase}[0]{\emph{OUT}}

\definecolor{hyperlinkblue}{HTML}{0000AA}
\hypersetup{citecolor=hyperlinkblue} %

\lstdefinestyle{mystyle}{
    breaklines=true,
    basicstyle=\scriptsize\ttfamily,
    numbers=none,
    language={},
    framextopmargin=0pt,
    framexbottommargin=0pt,
    breakindent=0pt,
    showspaces = false,
    keywordstyle=\bfseries,
    showstringspaces=false,
    columns=fullflexible,
    morekeywords={},
}

\newtcblisting{prompt}[2][]{
    arc=0pt, outer arc=1pt,
    width=.95\linewidth,
    left=3mm,
    top=0mm,
    bottom=0mm,
    title=#2, 
    colback=blue!5!white,
    colframe=blue!20!black,
    fonttitle=\bfseries,
    listing only, 
    listing options={style=mystyle,deletekeywords={}},
    breakable,
    #1
}

\usepackage[capitalize]{cleveref}

\makeatletter
\AddToHook{cmd/appendix/before}{\crefalias{section}{appendix}}
\AddToHook{cmd/appendix/before}{\crefalias{subsection}{appendix}}
\AddToHook{cmd/appendix/before}{\crefalias{subsubsection}{appendix}}
\makeatother

\crefformat{section}{\S#2#1#3}

\crefrangeformat{section}{\S#3#1#4\crefrangeconjunction\S#5#2#6}

\crefmultiformat{section}{\S#2#1#3}{\crefpairconjunction\S#2#1#3}{\crefmiddleconjunction\S#2#1#3}{\creflastconjunction\S#2#1#3}

\newcommand{\crefrangeconjunction}{--}

\crefname{listing}{Lst.}{listings}
\crefname{line}{Lin.}{Lin.}
\crefname{appendix}{App.}{App.}

\newcommand{\appref}[1]{%
	\ifbool{includeappendix}{\cref{#1}}{the appendix}%
}
\newcommand{\Appref}[1]{%
	\ifbool{includeappendix}{\cref{#1}}{The appendix}%
}

\title{\ours{}: Provable RAG Dataset Inference \\ via LLM Watermarks}

\author{Nikola Jovanovi\'c, Robin Staab, Maximilian Baader, Martin Vechev\\
ETH Zurich\hfil\\
\texttt{\{nikola.jovanovic, robin.staab, mbaader, martin.vechev\}@inf.ethz.ch}\\
}

\begin{document}

\maketitle

\vspace{-0.1in}
\begin{abstract}
RAG enables LLMs to easily incorporate external data, raising concerns for data owners regarding unauthorized usage of their content.
The challenge of detecting such unauthorized usage remains underexplored, with datasets and methods from adjacent fields being ill-suited for its study. 
We take several steps to bridge this gap. 
First, we formalize this problem as (black-box) RAG Dataset Inference (\emph{RAG-DI}). 
We then introduce a novel dataset designed for realistic benchmarking of RAG-DI methods, alongside a set of baselines. 
Finally, we propose \textsc{Ward}, a method for RAG-DI based on LLM watermarks that equips data owners with rigorous statistical guarantees regarding their dataset's misuse in RAG corpora. 
\textsc{Ward} consistently outperforms all baselines, achieving higher accuracy, superior query efficiency and robustness. 
Our work provides a foundation for future studies of RAG-DI and highlights LLM watermarks as a promising approach to this problem.
\end{abstract} 

\section{Introduction} \label{sec:introduction}

Retrieval-augmented generation (RAG) has emerged as a popular approach to mitigate limitations of large language models (LLMs) such as hallucinations, the high cost of adapting to new knowledge via fine-tuning, and the inability to back up claims by sources \citep{LewisPPPKGKLYR020}. 
By integrating retrieval, LLMs gain in-context access to large corpora of high-quality, up-to-date data, enabling them to generate more accurate and source-supported responses.
To maintain relevance, RAG providers must continuously update their corpus with new data.
However, this raises concerns regarding the unauthorized usage of documents, particularly when publicly available documents are used without the owner's permission \citep{nyt, WeiWJ24}{; see~\cref{app:motivation} for a more elaborate discussion of this issue and its prevalence in practice.}
Crucially, there is currently no way to conclusively prove such unauthorized usage by a RAG system, and enforce an opt-out by the  owner.

\begin{figure*}[t] 
    \centering
    \includegraphics[width=\textwidth]{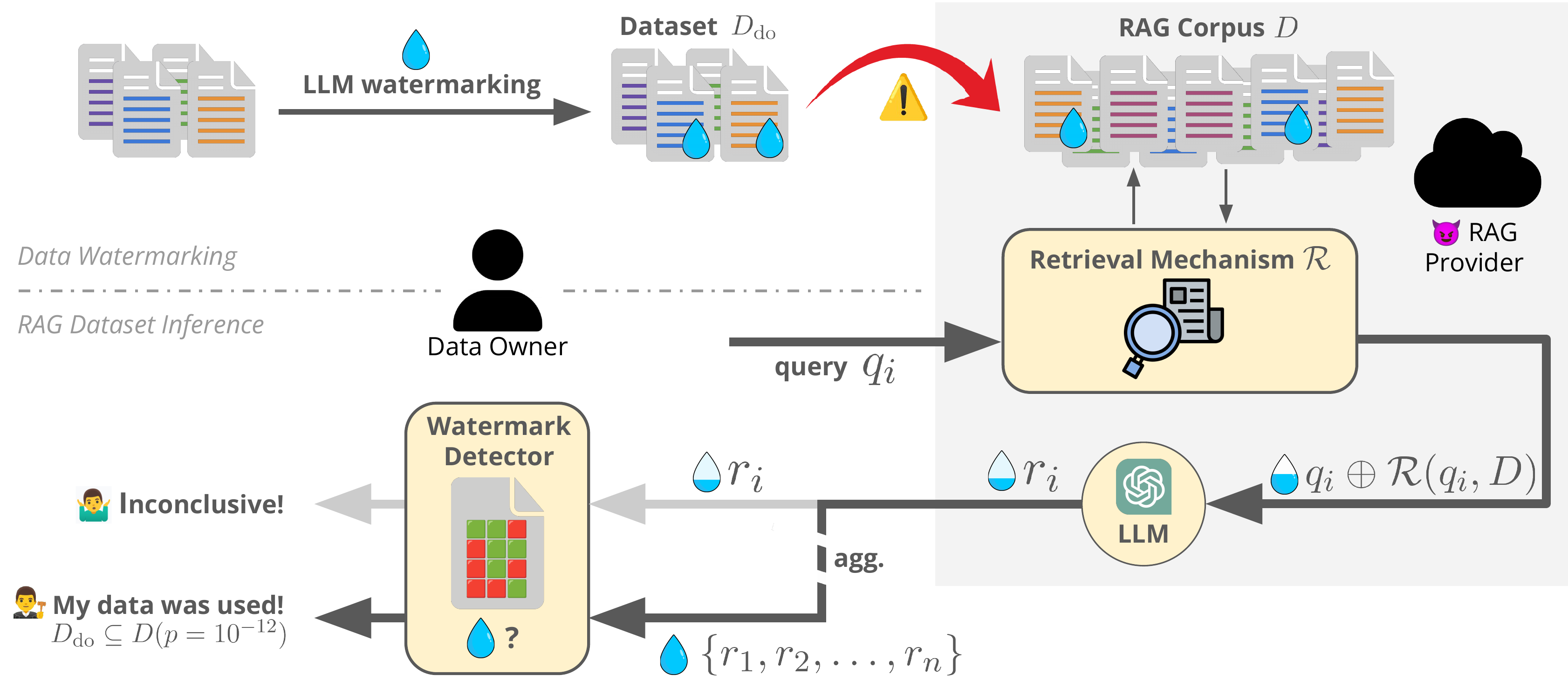}
    \caption{Overview of RAG Dataset Inference using \ours{}, our method based on LLM watermarks.}
    \label{fig:accept}
    \vspace{-0.15in}
\end{figure*}

\paragraph{RAG Dataset Inference (\problem{})}
We formalize this problem as \emph{RAG Dataset Inference} (\emph{\problem{}}), 
where a data owner aims to detect unauthorized inclusion of their dataset in a RAG corpus via black-box queries (\cref{fig:accept}).
In the first comprehensive study of this problem, we observe that existing datasets, used in adjacent works on RAG privacy, are not suitable for \problem{}.
First, the samples in these datasets may have been used in contemporary LLM training, complicating realistic evaluations where RAG corpora consist of new data.
Second, these datasets do not model \emph{fact redundancy}, a key property of real-world RAG, where multiple documents have similar content, either due to scraping data from various sources, \eg news~\citep{ragsurvey}, or due to chunking.
Another challenge in studying \problem{} stems from the lack of baselines applicable in a realistic black-box setting.

\paragraph{Foundations for \problem{}}
In this work, we take multiple steps to bridge these gaps: First, we introduce \newdata{}, a new dataset specifically designed for \problem{} evaluation under realistic conditions.
\newdata{} contains fictional articles that are by design not part of any LLM training data, and can enable evaluations under fact redundancy, enabling accurate assessment of \problem{} methods.
Second, we adapt prior work on RAG Membership Inference Attacks (MIAs)~\citep{sib, ibm} to the \problem{} problem, and propose a simple baseline~\facts{}.
In our evaluation on \newdata{}, we find that (i) despite its extreme simplicity, \facts{} outperforms other baselines in settings with no fact redundancy, further underscoring the drawbacks of existing datasets, and (ii) when fact redundancy is present, no baseline achieves satisfactory performance.
This highlights the need for novel approaches capable of reliably identifying unauthorized usage of documents in RAG corpora.

\paragraph{LLM Watermarks as a reliable \problem{} method}
To this end, we introduce \ours{}, a \problem{} method that protects the data owner's dataset by imprinting LLM watermarks~\citep{kgw,stanford}. 
As \cref{fig:accept} illustrates, given a limited number of black-box queries $q_i$ to the retrieval-augmented LLM, the data owner can detect even small traces of the watermark across responses $r_i$, and for the first time obtain rigorous statistical guarantees regarding the usage of their dataset in the RAG corpus, enabling them to effectively audit the RAG provider.
In our experiments on \newdata{}, we show that \ours{} consistently outperforms all baselines across a variety of challenging settings, showing high rate of true positives (often 100\%), with no false accusations.
Further, due to its robustness across settings, and the fact that it engages only in natural-looking interactions with the retrieval-augmented LLM, \ours{} retains its performance even under attempts by the RAG provider to prevent unintended uses of the system.
These results underscore the effectiveness of \ours{} in enforcing data usage policies and protecting data owners' rights in the context of RAG systems.

\paragraph{Main contributions} We make the following key contributions:
\begin{itemize}
    \item We formalize a previously unexplored problem, RAG Dataset Inference (\problem{}), where a data owner aims to detect unauthorized usage of their dataset in a RAG system (\cref{sec:setting}).
    \item We facilitate research on this problem by (i) proposing a new dataset \newdata{}, specifically designed for benchmarking \problem{} methods under realistic conditions, and (ii) introducing an initial set of baseline methods (\cref{ssec:setting:data} and \cref{ssec:setting:baselines}). 
    \item We propose LLM watermarks as a way to provably, robustly and reliably detect unauthorized data usage in RAG corpora, introducing \ours{} as a novel \problem{} method (\cref{sec:method}).
    \item Our experimental evaluation in a wide range of settings affirms the fundamental limitations of existing datasets and all \problem{} baselines, and demonstrates the effectiveness of \ours{}, which consistently shows high accuracy, query efficiency, and robustness (\cref{sec:evaluation}). Our source code and the \newdata{} dataset are publicly available at \url{https://github.com/eth-sri/ward}.
\end{itemize}  
\section{Background} \label{sec:background}

\paragraph{Retrieval-augmented generation (RAG)} 
RAG is a common way to enhance LLMs: For a given user query $q$, the $k \in \mathbb{N}$ most relevant documents $D_q = \mathcal{R}(q, D) \subseteq D$ are retrieved from a corpus $D$, using a retrieval method $\mathcal{R}$ \citep{ragsurvey}.
The query $q$ is generally combined with $D_q$, and fed into an LLM to generate a more factual response $r = \llm(q, D_q)$. Expanding $D$ also enables access to new information without costly retraining~\citep{LewisPPPKGKLYR020}.
RAG is especially suitable for domains where new information is generated often, \eg news articles or software documentation.
In this work we use $\mathcal{M}^\star$ to denote a \emph{RAG system}, \ie a retrieval-augmented LLM $\mathcal{M}$.

\paragraph{LLM watermarking}
LLM watermarks enable model owners to provably and reliably track text generated by their LLM. 
In this work, we focus on the prominent \emph{red-green watermarks} \citep{kgw}.
During text generation, at each step $t$, the token vocabulary $V$ is split into two parts, $\gamma |V|$ \emph{green} (encouraged) tokens, and $(1-\gamma) |V|$ \emph{red} (discouraged) tokens, for some $\gamma \in [0,1]$.
Given \emph{context width} $h$, the split is commonly a function of tokens at positions ${t-h}, \dots, t$, as well as a secret salt.
To add the watermark, the logits of green tokens are increased by $\delta \in \mathbb{R}_{\geq 0}$, boosting their probability of being sampled. 
The watermark is detected using a statistical test, based on the expectation that non-watermarked text of length $T$ has $\gamma T$ green tokens.
Namely, we use the z-score
\begin{equation} 
    z = \frac{|s|_g - \gamma T}{\sqrt{\gamma (1-\gamma) T}},
    \label{eq:zscore}
\end{equation}
where $|s|_g$ is the number of green tokens in a given text $s$, and $T=|s|$.
From here, we derive the $p$-value $p = 1 - \Phi(z)$, where $\Phi$ is the CDF of the standard normal distribution, and consider the text watermarked if $p<\alpha$ for some threshold $\alpha$.
While not perfectly robust, these watermarks generally persist under moderate text transformations, such as paraphrasing or segment omission~\citep{markmy,kgw2,radioactivity}, making them suitable for our setting, where our goal will be to propagate the watermark signal through the RAG pipeline, as we will describe in~\cref{sec:method}.

\paragraph{RAG membership inference attacks}
While RAG is a relatively novel concept, recent work already studies membership inference attacks (MIAs) in this setting,
 proposing two methods that we denote \sib{}~\citep{sib} and \ibm{}~\citep{ibm}.
A MIA's goal is to output  $\mathrm{mi}(d, \ragsys)=1$ if a document $d$ is part of the retrieval corpus $D$ of a RAG system $\ragsys$, or $\mathrm{mi}(d, \ragsys)=0$ otherwise, based only on queries to $\ragsys$.
To this end, \sib{} queries $\ragsys$ with $q$, a prefix of $d$, to obtain the response $r = \llm(q, D_q)$.
Then, it computes two scores: the cosine similarity between the embeddings of $d$ and $r$, and the perplexity of $r$, outputting $\mathrm{mi}(d, \ragsys)=1$ if the similarity is above a threshold $\theta_\text{similarity}$, and the perplexity is below a threshold $\theta_\text{perplexity}$, both trainable parameters. 
Note that this requires gray-box access to $\ragsys$ for perplexity computation. 
The other method, \ibm{}, directly prompts $\ragsys$ to answer if $d$ is in the context. 
If it replies positively, they set $\mathrm{mi}(d, \llm)=1$.
In~\cref{sec:setting}, we will introduce the black-box RAG dataset inference setting, and adapt both baselines to it.
Notably, while dataset inference was studied alongside MIA for training data (see~\cref{sec:related_work}), no prior work studies it for RAG.

\section{RAG Dataset Inference} \label{sec:setting}
We formalize the problem of \emph{{RAG Dataset Inference}} (\emph{\problem{}}), and present our contributions aimed at facilitating studies of this problem. 
In~\cref{ssec:setting:data}, we make the case that existing datasets commonly used for adjacent tasks (e.g., RAG MIA) are fundamentally unsuitable for \problem{}, and propose a new dataset in an attempt to address these shortcomings. 
In~\cref{ssec:setting:baselines}, we establish a set of baselines for \problem{}, by adapting RAG MIA work introduced in~\cref{sec:background}, and proposing a simple baseline, \facts{}. 

\paragraph{The \problem{} problem}
The key entities in \problem{} are the \emph{data owner}, who aims to protect their $n$-document dataset $\ownerdata$ from unauthorized usage in a RAG corpus, and the \emph{RAG provider}, who exposes black-box access to their retrieval-augmented LLM $\ragsys$, which uses a corpus $D$.
The data owner's goal is to determine if $\ownerdata \subseteq D$, i.e., whether their data was secretly included in the corpus. 
To this end, they may proactively modify $\ownerdata$ before publishing it, and can query $\ragsys$ in a black-box way, aiming to minimize the number of such queries.
Crucially, the data owner makes a single dataset-level decision, as opposed to document-level decisions of MIAs.
Formally, a \problem{} method $di$ should output $\mathrm{di}(\ownerdata, \ragsys)=1$ if $\ownerdata \subseteq D$ (\incase{} case) and $0$ otherwise (\outcase{} case).

\subsection{A dataset suitable for \problem} \label{ssec:setting:data}

To enable suitable evaluation of \problem{} methods, we require a dataset of documents with the following properties.
First, we aim to match a key use-case of RAG (as described in~\cref{sec:background}) where up-to-date knowledge is added to $D$ instead of costly repeated fine-tuning of $\llm$.
To model this, our documents should provably not be part of the training data of $\llm$, i.e., of current open/closed LLMs, as those will be used to instantiate $\llm$ when studying \problem{}.
Second, to model the practical case where knowledge is redundant and spread across multiple sources (e.g., news articles, a common motivating example for RAG~\citep{ragsurvey}), the dataset should contain documents with overlapping topics and information (\emph{fact redundancy}).
As we will empirically demonstrate in~\cref{sec:evaluation}, while fact redundancy is more realistic, it makes \problem{} significantly harder.

\begin{wrapfigure}[31]{r}{0.38\textwidth}
    \centering
    \includegraphics[width=\linewidth]{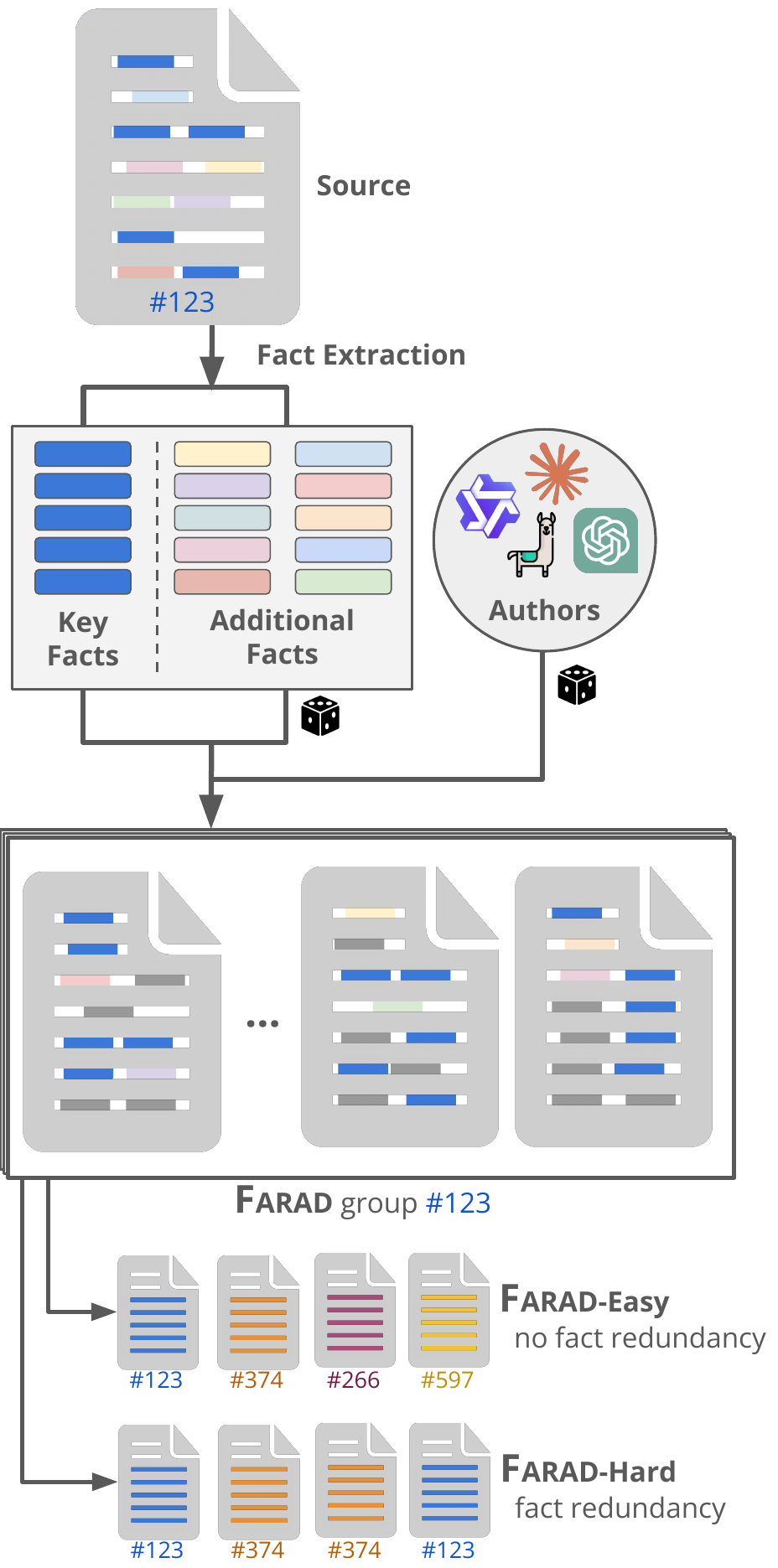}
    \caption{Overview of the generation pipeline of \newdata{}, and the resulting \easy{} and \hard{} evaluation settings.}
    \label{fig:pipeline}
\end{wrapfigure}   
\paragraph{Current state}
As there is no prior work on \problem{}, we turn to related work on RAG privacy~\citep{ibm,sib,goodbad}, including MIAs introduced in~\cref{sec:background}.
We observe that evaluations in these works rely primarily on \emph{EnronEmails}~\citep{enron} and \emph{HealthcareMagic} datasets~\citep{meddialog,healthcaremagic}, motivated by the presence of PII in their samples.  
However, it can not be ruled out that these datasets were used to train contemporary LLMs, as e.g., \emph{EnronEmails} has been publicly available since 2004. 
More importantly, fact redundancy is by design not satisfied in either of these cases.
These shortcomings motivate us to construct a new dataset, \newdata{} (\emph{Fact-Redundant Article Dataset}), tailored to \problem{}.

\paragraph{The \newdata{} dataset}
\newdata{} consists of a number of \emph{groups}.
Each group contains articles that share a topic and a significant amount of information, but are independently written by a different (LLM) author. 
As our data source we use \repliqa{}~\citep{repliqa}, that contains articles about fictional entities and events, which by design ensures that this knowledge was not present in any LLM training data.
\repliqa{} is released gradually---to create \newdata{}, we use split $0$ as the only one available at the time of writing, but plan to expand this to future splits.
 
Each \repliqa{} article is a \emph{source} for one of our groups---we use the pipeline illustrated in~\cref{fig:pipeline}.
First, we prompt \gptBig{} to distill the information content of the article into 5 self-contained \emph{key facts}, crucial to understanding the article, and 10 self-contained \emph{additional facts}, that are present in the article but not essential to its key message (see~\cref{app:prompts:data}).
Next, to create a group, we sample an author model from the set $\mathcal{A}= \{$\gptBig{},~ \sonnet{},~ \llamaBig{},~ \qwen{}$\}$ of state-of-the-art LLMs.
The selected model writes a $500$--$1000$ word article that must include all $5$ key facts, $2$ randomly sampled additional facts, and is encouraged to invent additional quotes, hypotheses, or personal opinions, as long as they do not contradict any present fact.
The diversity of authors, sampled facts, and (desirable) hallucinated content results in a varied set of fictional articles grounded in the same core knowledge, which satisfies our earlier requirements.
While this pipeline can be extended to larger/more groups, more authors, and different fact combinations, we limit \newdata{} to $3591$ groups ($3391$ for testing, and $200$ for training) of $4$ articles each, one per author from $\mathcal{A}$.

\paragraph{Use in \problem{} evaluation}
In our experiments in \cref{sec:evaluation}, we use \newdata{} to create two evaluation settings: \easy{} and \hard{}, illustrated in~\cref{fig:pipeline} (bottom).
The \easy{} setting follows the traditional setup described above where facts are uniquely represented, by always using at most $1$ article per \newdata{} group. 
The \hard{} setting introduces fact redundancy by always including all $4$ articles from a group. 
As we demonstrate shortly, the \easy{} setting is solvable by a simple baseline, while the \hard{} setting, as the realistic one and our main focus, is a significant challenge for all baseline methods.
{In~\cref{app:limitations:data} we discuss potential improvements to our data generation pipeline as items for future work.}

\subsection{RAG-DI Baselines} \label{ssec:setting:baselines}
We proceed to establish an initial set of baselines for the \problem{} problem. For this we adapt existing RAG MIAs, \ibm{} and \sib{}, introduced in \cref{sec:background}, and additionally propose a simple baseline, \facts{}.

\paragraph{Adapting RAG MIAs}
By design, the existing RAG MIAs make document-level decisions, i.e., they decide if a \emph{single document} is in the RAG corpus ($\mathrm{mi}(d, \ragsys)=1$) or not ($\mathrm{mi}(d, \ragsys)=0$).
We extend this to dataset-level decisions of \problem{} in an (empirically) optimal way.
First, we apply a given method to each $d \in \ownerdata$ to obtain $n$ binary decisions, and set $s(\ownerdata, \ragsys) = \frac{1}{n} \sum_{d \in \ownerdata} \mathrm{mi}(d, \ragsys)$. 
We then instantiate several \incase{} and \outcase{} cases of \problem{} using the training set, and compute the corresponding $s_{\text{in}}$ and $s_{\text{out}}$.
We finally make a dataset-level statement $\mathrm{di}(\ownerdata, \ragsys) = 1$ if $s(\ownerdata, \ragsys) > \frac{1}{2} (s_{\text{in}} + s_{\text{out}})$ and $\mathrm{di}(\ownerdata, \ragsys) = 0$ otherwise.
During this process, for \sib{}, we also grid search over its trainable parameters $\theta_{\text{similarity}}$ and $\theta_{\text{perplexity}}$ (as in~\citet{sib}), choosing values that maximize $s_{\text{in}} - s_{\text{out}}$.
We note that \sib{} is proposed as a gray-box method; we adapt it to our black-box setting by using an auxiliary language model to estimate the perplexity of responses.

\paragraph{The \facts{} baseline}
To substantiate our point about the effect of fact multiplicity on the hardness of \problem{}, we introduce a simple baseline, \facts{}.
As the above RAG MIAs, \facts{} is a document-level method.
Given a document $d$, \facts{} prompts an auxiliary LLM to generate a single question that is only answerable by reading $d$. 
Then, $\ragsys$ is prompted with that question, and if it deems it unanswerable, we set $\mathrm{mi}(d, \ragsys) = 0$, and set it to $1$ otherwise.
We aggregate such document-level decisions into a corpus-level decision by fitting a threshold on the training set, as described above. 
In~\cref{sec:evaluation}, we will demonstrate that this simple method is sufficient to solve the \easy{} setting more reliably than other baselines, yet is, due to fact multiplicity, extremely unreliable in the \hard{} setting.
\section{LLM Watermarking as an Effective \problem{} Method} \label{sec:method}

Before describing our proposed \problem{} method based on LLM watermarks, we first outline three key design requirements that a desirable \problem{} method should fulfill. 
We require the following:
\begin{enumerate}
    \item \textit{Monotonicity}. With more queries to the retrieval-augmented LLM $\ragsys{}$, the accuracy of the method's predictions should consistently improve, preferably at a high rate. 
    \item \textit{Guarantees}. The method should be able to provide a statistical guarantee for its decision, with exceedingly rare and well-controlled Type 1 errors, as falsely accusing RAG providers is highly undesirable in practice, and undermines the trust in the method.
    \item \textit{Robustness}. The method should maintain high accuracy under diverse evaluation settings, including attempts by the RAG provider to actively conceal unauthorized data usage.
\end{enumerate}
In~\cref{app:motivation} we motivate these requirements in detail by reflecting on the context around \problem{}.
As we demonstrate in~\cref{sec:evaluation}, all \problem{} baselines introduced in~\cref{ssec:setting:baselines} violate \emph{all} requirements to some extent. 
To address this, we propose \ours{} (\emph{Watermarking for \problem{}}), a proactive \problem{} method that is based on LLM watermarks, and discuss why it is likely to fulfill all stated desiderata.

\paragraph{\ours{}: \problem{} via LLM watermarking}
We assume the data owner has protected each $d_i \in \ownerdata$ by embedding an LLM watermark either via a human-in-the-loop procedure or (as in this work) by rephrasing each document with a watermarked LM.
While, in principle, any LLM watermark can be applied, we focus on popular red-green watermarks (see~\cref{sec:background}).
In~\cref{ssec:eval:quality}, we confirm that this results in quality texts, faithful to the original ones.
To audit the RAG provider's corpus, for each $d_i \in \ownerdata$, \ours{} generates an open-ended content-related question $q_i$, and queries $\ragsys{}$.
If $d_i \in D$ (\incase{} case), we expect the retrieval method $\mathcal{R}$ to introduce watermarked content from $d_i$ into the LLM context.
As noted in~\cref{sec:background} and validated in \cref{sec:evaluation}, the robustness of watermarks to text transformations is then sufficient to propagate the traces of the signal to $r_i = \ragsys{}(q_i)$, the final response of the LLM.

\paragraph{Boosting a weak signal}
Requiring that each $r_i$ is flagged as watermarked, i.e., has watermark detector p-value $p < \alpha$, would be a strong assumption, as the watermark signal is likely to degrade throughout the RAG pipeline.
However, this is not necessary for \ours{} to be effective.
Instead, following~\citet{radioactivity}, after $n$ queries we compute a \emph{joint} p-value $R = \{r_1, \ldots, r_n\}$, directly corresponding to the null hypothesis \emph{``the data owner's dataset $\ownerdata$ is not in the RAG corpus $D$''}.

This joint p-value can satisfy $p < \alpha$, i.e., reject the null hypothesis, even when individual $r_i$ carry only weak watermark signal, that would individually not reject it. 
To illustrate this, given a desired p-value threshold $\alpha$, \cref{eq:zscore} implies that the required ratio of green tokens in $R$ is at least 
\begin{equation} 
    \gamma' \geq \Phi^{-1}(1-\alpha) \cdot \sqrt{\gamma (1-\gamma) / |R|^{\oplus}} + \gamma, \label{eq:scaling} 
\end{equation}
\begin{wrapfigure}{R}{0.35\textwidth}
    \vspace{-0.08in}
    \centering  
    \includegraphics[width=0.33\textwidth]{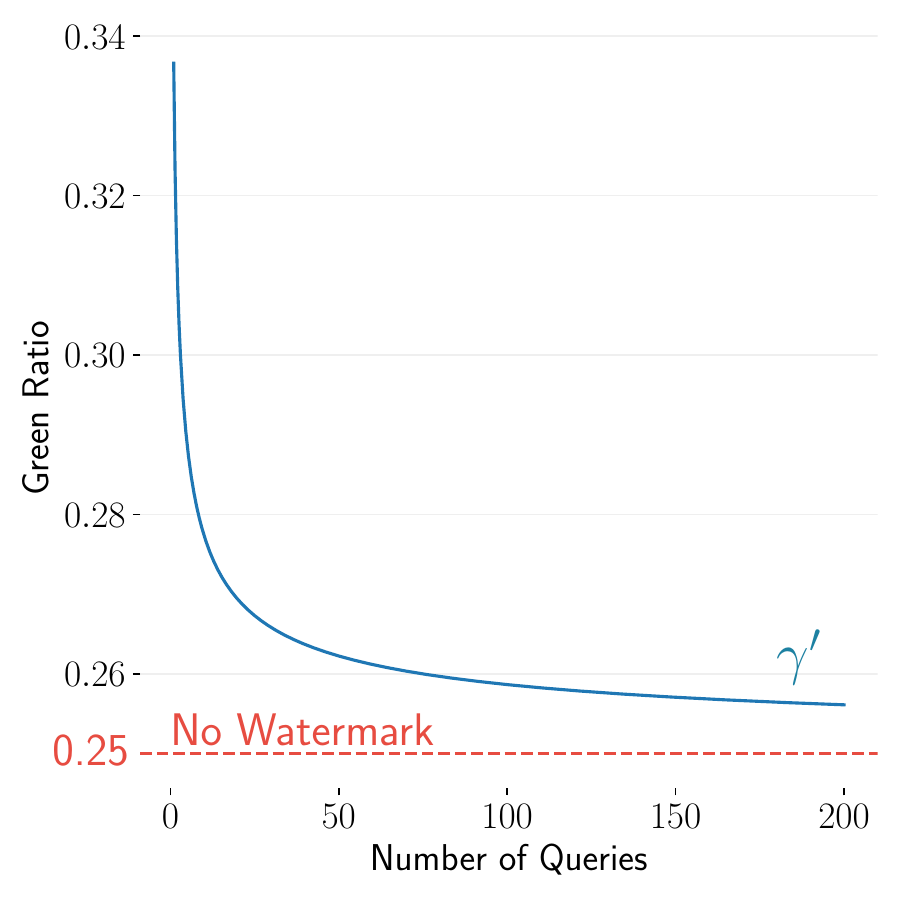}
    \vskip -0.3em
    \caption{Green ratio to pass watermark detection with given \#queries.}
    \label{fig:scaling}
    \vspace{-25pt}
\end{wrapfigure}  

where $\Phi$ is the standard normal distribution CDF, and $|R|^{\oplus}$ the total length of responses in $R$. 
This lower bound decreases quickly as $|R|^{\oplus}$ increases: In~\cref{fig:scaling} we plot the lower bound as a function of $n$, assuming $\forall i\colon~|r_i|=400$ tokens, \mbox{$\alpha \approx 3 \cdot 10^{-5}$} (z-score of at least $4$), and $\gamma=0.25$, as in our experiments.
For $n \geq 100$, if propagation of the watermark through the RAG pipeline increases the ratio of green tokens by only $1\%$, it is already detectable with high confidence.
This makes \ours{} viable, satisfying the \emph{Guarantees} requirement, unlike any baseline. 
It also contributes to \emph{Monotonicity}: assuming each ${d}_i$ has a green token ratio of $\gamma''$, which propagated through $\ragsys{}$ reduces to $\gamma' \in (\gamma, \gamma'')$, the p-value for \incase{} cases strictly decreases for more queries.
In~\cref{ssec:eval:desiderata} we experimentally confirm that \ours{} scales consistently and efficiently in the number of queries. 

\paragraph{Test validity and signal leakage}

Another requirement for \emph{Monotonicity} is that the p-value for \outcase{} cases is well-calibrated, ensuring exceedingly rare false positives.
As pointed out in prior work~\citep{kgw,radioactivity}, to ensure independence between tokens scored by the watermark detector, it is necessary to ignore duplicate $h$-grams across $R$.
While this reduces the number of useful tokens per query, our results in~\cref{sec:evaluation} show that it does not affect \ours{} performance.
 
For our statistical test to be valid, we must ensure that the watermark signal can only originate from the use of $\ownerdata$ in $D$, i.e., that it is not otherwise \emph{leaked} to the responses $r_i$.
In particular, we observed that watermarked queries, or queries based on watermarked versions of $d_i$, drastically increase false positives.
To confirm that \ours{} does not suffer from this issue, in~\cref{ssec:eval:desiderata}, we verify that the p-values in \outcase{} cases are distributed in $[0,1]$ roughly as expected with no noticeable decrease in $|R|$.

\paragraph{Scheme choice}  
The key parameter of red-green schemes is $h$, the context width. 
In common applications of LLM watermarks, a low $h$ is not recommended, as it makes the watermark easy to steal via repeated queries~\citep{kgw2,ws}. 
In the context of \problem{}, this is less of a concern, as the data owner never exposes unconstrained query access to watermarked content, as is the case when watermarking LLMs.
Thus, as low $h$ benefits watermark propagation through RAG, we use $h=2$.
This contributes to high \emph{Robustness} of \ours{}, empirically validated in~\cref{ssec:eval:desiderata}.
We discuss other scheme parameters and present ablation studies over each in \cref{ssec:eval:ablations}.

\paragraph{Practicality}
Another aspect making \ours{} robust is that it is inconspicuous, as it operates on natural-looking documents, queries, and responses (confirmed in~\cref{ssec:eval:add_considerations}), through legitimate use of the RAG system---this makes potential attempts by the RAG provider to thwart malicious interactions less effective. 
This differs from baselines such as \ibm{}, which directly reveal the intention to leak the information about $D$.
As a final advantage, we note that, in contrast to all baselines, \ours{} does not require any training or adaptation to \problem{}, as LLM watermarks naturally apply to this task.
\section{Experimental Evaluation} \label{sec:evaluation}

We evaluate the \problem{} baselines (\cref{ssec:setting:baselines}) and \ours{} (\cref{sec:method}) on the \newdata{} dataset (\cref{ssec:setting:data}).
\cref{ssec:eval:main_results} presents our main experiment.
In~\cref{ssec:eval:desiderata} we focus on desiderata from \cref{sec:method}, showing that only \ours{} does not violate them. 
In \cref{ssec:eval:add_considerations} we validate several key assumptions, and in~\cref{ssec:eval:ablations} present additional ablations.
 
\paragraph{Setup} 
Our experimental setup follows \cref{sec:setting}: we use \newdata{} to define two evaluation settings, and in both evaluate \incase{} and \outcase{} cases, i.e., where the data owner's data \emph{is} (resp. \emph{is not}) contained in $D$.
We use $|\ownerdata|=200$, and $|D|=800$ for \newdatae{}, and $|D|=3000$ for \newdatah{} (sampling detailed in ~\cref{app:expdetails}).
We note that \ours{} only depends on $|\ownerdata{}|$, but not $|D|$.
We use several LLMs as $\mathcal{M}$: \gptSmall, \haiku, and \llamaMid, and vary the system prompt: we use a short \emph{naive} prompt (\naive{}) with basic RAG instructions, and a longer \emph{defense} (\defended{}) prompt, which models a RAG provider that instructs the model {to not regurgitate sources verbatim, and refuse attempts to learn about the exact LLM context (see \cref{app:prompts:rag})}.
Each experiment is run with $5$ random seeds.
To ensure a controlled setting, in our main experiments, we assume a perfect retrieval system that always {retrieves the most relevant documents (see~\cref{app:retrieval} for a detailed explanation)}. 
In \cref{ssec:eval:add_considerations}, we show that \ours{} works equally well on a practical end-to-end RAG. 
If not specified otherwise (see \cref{ssec:eval:ablations}), the RAG uses $k=3$ shots.
For {methods that utilize auxiliary LMs (including \ours{}, which starts by watermarking \ownerdata{})} we use \llamaSmall{}, and to compute the cosine similarity in \sib{} we use paragraph-level \sentenceTf{} with BERTScore aggregation~\citep{bertscore}.
For \ours{}, we use \textit{PositionPRF}~\citep{kgw2}, $h=2$, and $\delta=3.5$, ablating these in \cref{sssec:eval:ablation_parameters}. 
We list all prompts, and examples of \newdata{} samples and watermarked documents, in \cref{app:prompts}.

\subsection{Main Results}
\label{ssec:eval:main_results}

\begin{figure}[t]
    \centering
    \includegraphics[width=\textwidth]{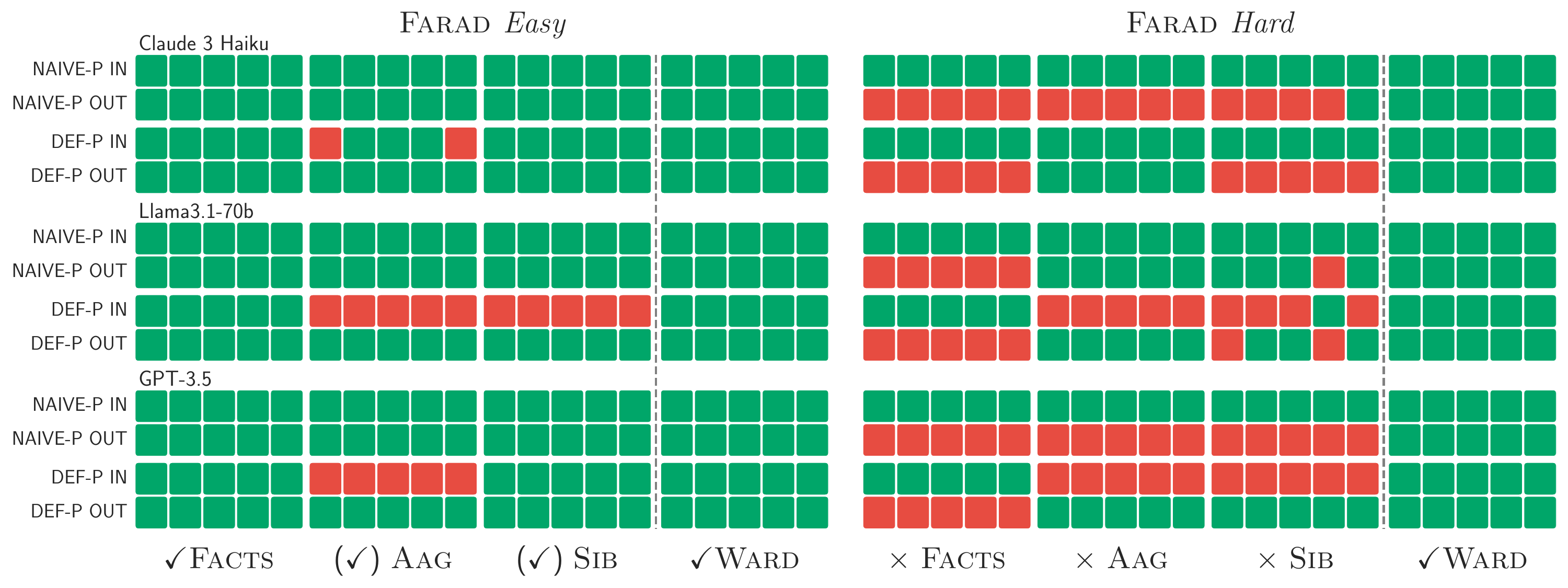}
    \caption{Evaluation of all methods on \newdata{} in both \easy{} and \hard{} settings, and with both \naive{} and \defended{} system prompts. We run each method with $5$ random seeds, resulting in $5$ squares. A red square indicates a false negative in the \incase{} case, and a false positive in the \outcase{} case. All methods perform well in the \easy{} setting, while only \ours{} consistently performs well in the \hard{} setting.}
    \label{fig:main_exp}
    \vskip -1em
\end{figure}

We present our main results in \cref{fig:main_exp}, where we evaluate all \problem{} baselines and \ours{} across several settings, models, prompts, and random seeds.
We make several key observations.

First, all baselines perform somewhat well in the \easy{} setting (no fact redundancy).
However, our extremely simple \facts{} baseline obtains perfect results, outperforming both \ibm{} and \sib{}.
This emphasizes that traditional non-redundant datasets (see~\cref{ssec:setting:data}) fail to capture the complexity of \problem{}, and can provide an incomplete view of the capabilities of \problem{} methods. 
We also observe that already in the \easy{} setting, straightforward system prompt defenses significantly impact both \ibm{} and \sib{}, leading to a noticeable increase in false negatives. 
We further investigate defenses in \cref{ssec:eval:desiderata}.

In the \hard{} setting, all baselines fail to perform consistently, inducing both false positives and negatives.
This can be attributed to their shortcomings in handling fact redundancy: \facts{} directly relies on facts, while \sib{} and \ibm{} rely on semantic similarity influenced by factual content.
Notably, only \ours{} achieves 100\% accuracy across all settings, models, and system prompts, showing that despite the retrieval of documents with partially overlapping facts, watermarking provides a reliable signal for dataset inference.
This backs up our claims regarding the importance of fact-redundancy for realistic \problem{} evaluation and highlights the potential of watermarking as an approach to \problem{}.

\subsection{Desiderata}
\label{ssec:eval:desiderata}
We next demonstrate how baselines violate the desiderata from \cref{sec:method}, which is not the case for \ours{}.

\paragraph{Monotonicity}
 
\begin{figure}[t]
    \centering
    \begin{minipage}{0.48\linewidth}
      \includegraphics[width=\textwidth]{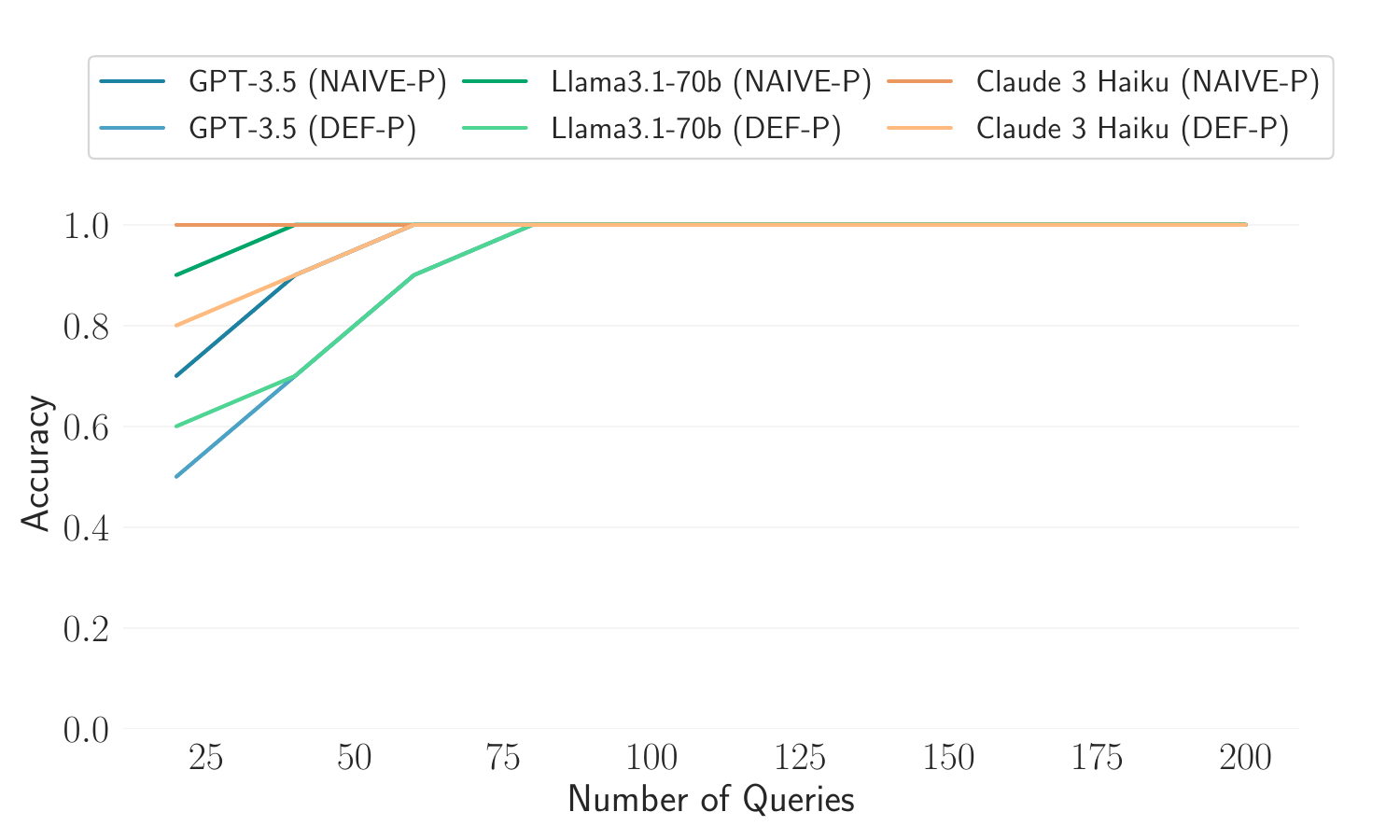}
    \end{minipage}
    \hfill
    \begin{minipage}{0.48\linewidth}
      \includegraphics[width=\textwidth]{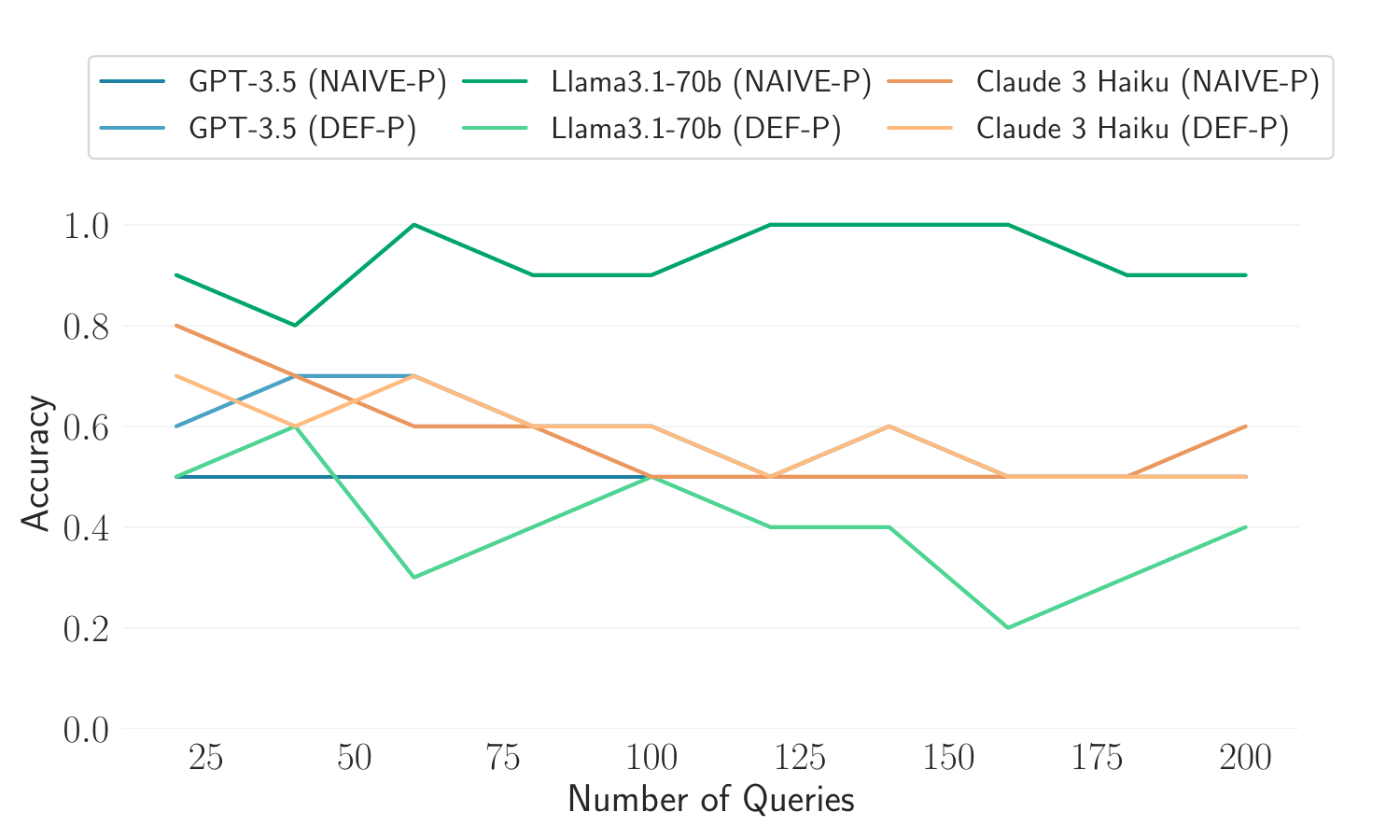}
    \end{minipage}
    \vskip -0.8em
    \caption{\ours{} (left) and \sib{} (right) accuracy as a function of $|\ownerdata{}|$ (i.e., the number of queries) in the \hard{} setting. \ours{} consistently improves, while \sib{} suffers from high variance in accuracy.}

    \label{fig:consistency}
\end{figure}

\begin{wrapfigure}{R}{0.48\textwidth}
  \centering
  \vspace{-0.5em}
  \begin{minipage}{\linewidth}
    \includegraphics[width=\textwidth]{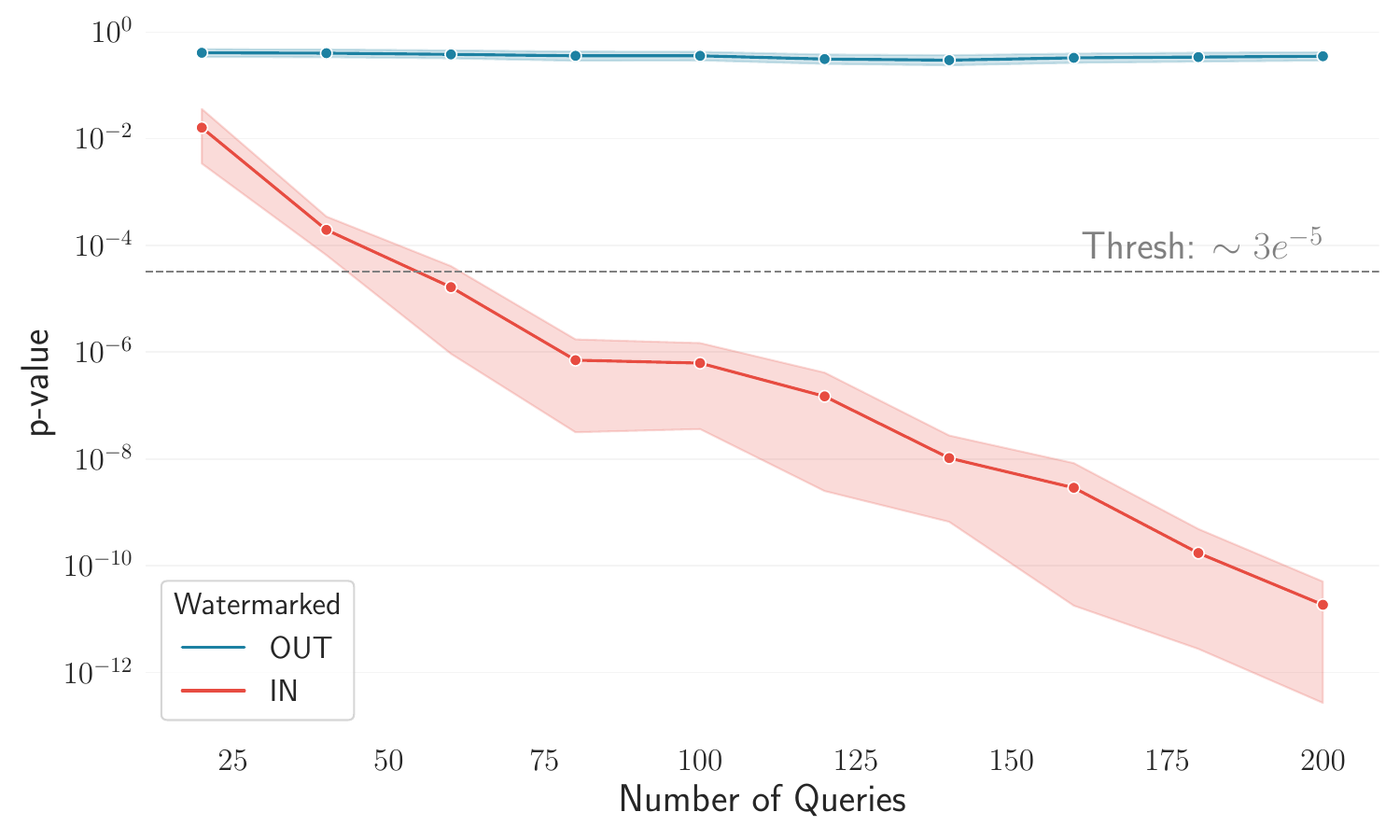}
    \caption{P-values as a function of \#queries in \incase{} (red) and \outcase{} (blue) cases.}
    \label{fig:nul_scaling}
  \end{minipage}
  \vskip -1.5em
\end{wrapfigure}

As stated in \cref{sec:method}, \problem{} methods should steadily improve with more queries, i.e., $|\ownerdata{}|$ for \ours{} (see \cref{ssec:eval:ablations} for a generalization).
We evaluate this by setting $|\ownerdata{}| \in \{20, 40, \ldots, 200\}$ in \newdatah{}, presenting the results of \ours{} and \sib{} in \cref{fig:consistency}.
\ours{} improves consistently with $|\ownerdata{}|$, reaching perfect accuracy across all settings for at most $80$ documents.
{
In our extended discussion of efficiency in~\cref{app:efficiency} we show that this translates to inexpensive API costs of below $\$1$, for all closed-source LLMs we consider.}
In contrast, \sib{}, besides never reaching full accuracy, exhibits strongly varying accuracy over $|\ownerdata{}|$, often decreasing despite using more queries. 
This is a consequence of the need to adapt the MIA baselines to the \problem{} setting---we find similar behavior across all baselines, and present an additional study of their decision thresholds in \cref{app:more_results:baselines}.

\paragraph{Guarantees}
In contrast to \emph{all} baselines, \ours{} inherits the guarantees of the watermarking scheme, directly providing a p-value for each decision. 
In~\cref{fig:nul_scaling}, we see that p-values rapidly decrease with more queries for \incase{}, but stay consistently close to the expected value for \outcase{}. 
Further, in \cref{tab:p_vals}, we show the \emph{max} p-values for \incase{} and \emph{min} p-values for \outcase{} in our main experiment, i.e., closest to a false negative/positive. All p-values are orders of magnitude from our decision boundary of $\approx 3e^{-5}$ (z-score of $4$), highlighting robustness to Type 1 errors and the fact that LLM watermark signals persist through a RAG pipeline.

\begin{table}[t]
  \centering
  \vskip -0.5em
  \caption{The p-values of \ours{}. We report the max p-values for \incase{} and min p-values for \outcase{} cases.}
  \vskip -0.7em
  \label{tab:p_vals}
  \resizebox{\textwidth}{!}{%
  \begin{tabular}{llcccccccccccc}
      \toprule
      & & \multicolumn{4}{c}{Claude 3 Haiku} 
        & \multicolumn{4}{c}{GPT-3.5}
        & \multicolumn{4}{c}{Llama3.1-70b} \\
      \cmidrule(lr){3-6} \cmidrule(lr){7-10} \cmidrule(lr){11-14}
      Data & Agg.
        & \multicolumn{2}{c}{DEF-P} 
        & \multicolumn{2}{c}{NAIVE-P} 
        & \multicolumn{2}{c}{DEF-P} 
        & \multicolumn{2}{c}{NAIVE-P} 
        & \multicolumn{2}{c}{DEF-P} 
        & \multicolumn{2}{c}{NAIVE-P} \\
      \cmidrule(lr){3-4} \cmidrule(lr){5-6} 
      \cmidrule(lr){7-8} \cmidrule(lr){9-10} 
      \cmidrule(lr){11-12} \cmidrule(lr){13-14}
      & & \outcase{} & \incase{} 
        & \outcase{} & \incase{} 
        & \outcase{} & \incase{} 
        & \outcase{} & \incase{} 
        & \outcase{} & \incase{} 
        & \outcase{} & \incase{} \\
      \midrule
      \multirow{2}{*}{\shortstack{\textsc{Farad}\\\emph{Easy}}} 
        & min 
        & $1.90e^{-01}$ & $2.07e^{-267}$ 
        & $1.38e^{-03}$ & $3.47e^{-301}$ 
        & $3.28e^{-01}$ & $9.53e^{-33}$ 
        & $2.13e^{-01}$ & $1.78e^{-78}$ 
        & $7.54e^{-03}$ & $6.20e^{-179}$ 
        & $1.32e^{-01}$ & $0$ \\
      & max 
        & $6.83e^{-01}$ & $1.57e^{-186}$ 
        & $2.83e^{-01}$ & $2.19e^{-253}$ 
        & $7.72e^{-01}$ & $1.54e^{-27}$ 
        & $9.17e^{-01}$ & $6.56e^{-60}$ 
        & $9.55e^{-01}$ & $1.52e^{-138}$ 
        & $5.05e^{-01}$ & $0$ \\
      \multirow{2}{*}{\shortstack{\textsc{Farad}\\\emph{Hard}}} 
        & min 
        & $6.03e^{-02}$ & $1.35e^{-48}$ 
        & $3.82e^{-01}$ & $5.08e^{-74}$ 
        & $2.04e^{-03}$ & $1.12e^{-12}$ 
        & $2.69e^{-01}$ & $1.53e^{-22}$ 
        & $1.18e^{-02}$ & $3.94e^{-24}$ 
        & $6.12e^{-02}$ & $1.22e^{-122}$ \\
      & max 
        & $3.99e^{-01}$ & $2.15e^{-25}$ 
        & $7.34e^{-01}$ & $3.16e^{-51}$ 
        & $7.80e^{-01}$ & $8.44e^{-10}$ 
        & $7.92e^{-01}$ & $1.31e^{-15}$ 
        & $7.47e^{-01}$ & $1.10e^{-11}$ 
        & $3.70e^{-01}$ & $2.42e^{-96}$ \\
      \bottomrule
  \end{tabular}%
  } 
  \vskip -1.3em
\end{table}

\begin{wrapfigure}[12]{R}{0.45\textwidth}
  \centering
  \vskip -1.5em
  \begin{minipage}{\linewidth}
    \includegraphics[width=\textwidth]{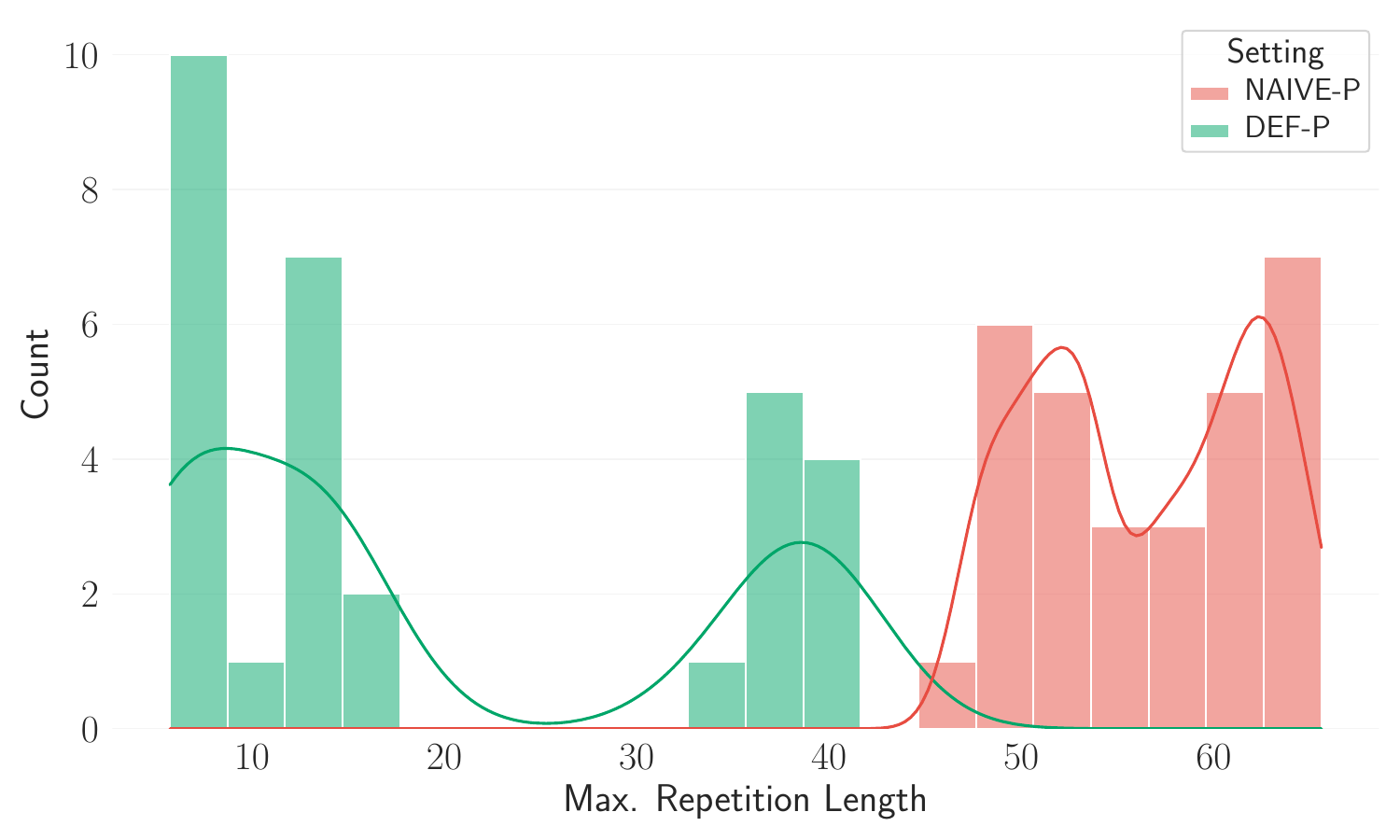}
    \caption{The effect of the \defended{} sysprompt.}
    \label{fig:maxrep}
  \end{minipage}
\end{wrapfigure}

\paragraph{Robustness}
Finally, we evaluate the methods' robustness to different settings. 
In particular, we further examine the \defended{} system prompt setting, shown in~\cref{fig:main_exp} to often prevent all baselines from obtaining any useful estimates. 
Notably, both \ibm{} and \sib{} rely on the model's \emph{willingness} to leak information about its context, either explicitly or by regurgitating similar content. 
To validate the effect of \defended{}, we show the longest token-string overlap between the response of $\ragsys{}$ and the \sib{} target document (\cref{fig:maxrep}), observing strictly less overlap with \defended{}, inducing false negatives.
Importantly, this does not prevent watermarks from propagating through the RAG pipeline, resulting in a reliable signal for \ours{}.

We additionally investigate a defense based on \textsc{MemFree} decoding~\citep{memfree}.
Namely, we adapt the decoding of $\ragsys{}$ to strictly prevent $n$-gram overlap with the retrieved documents (using $n=10$ as in~\citet{memfree}), modeling an even stronger attempt by the provider to protect the RAG corpus.
Our results (\cref{app:more_results:memfree}), show that even in this setting, \ours{} achieves full accuracy.
{We further discuss the potential for more elaborate defenses against \ours{} in~\cref{app:limitations:countermeasures}.}

\subsection{Additional Considerations}
\label{ssec:eval:add_considerations}
\paragraph{Modeling retrieval}

So far, our experiments have assumed an idealized case of perfect retrieval. 
We now justify this choice by running \ours{} on an end-to-end RAG system which uses OpenAI's \texttt{text-embedding-3-large} document embeddings, with $k=3$ and cosine similarity metric.

\begin{wrapfigure}[14]{R}{0.51\textwidth}
  \centering  
  \vskip -0.2em
  \begin{minipage}{\linewidth}
    \includegraphics[width=\textwidth]{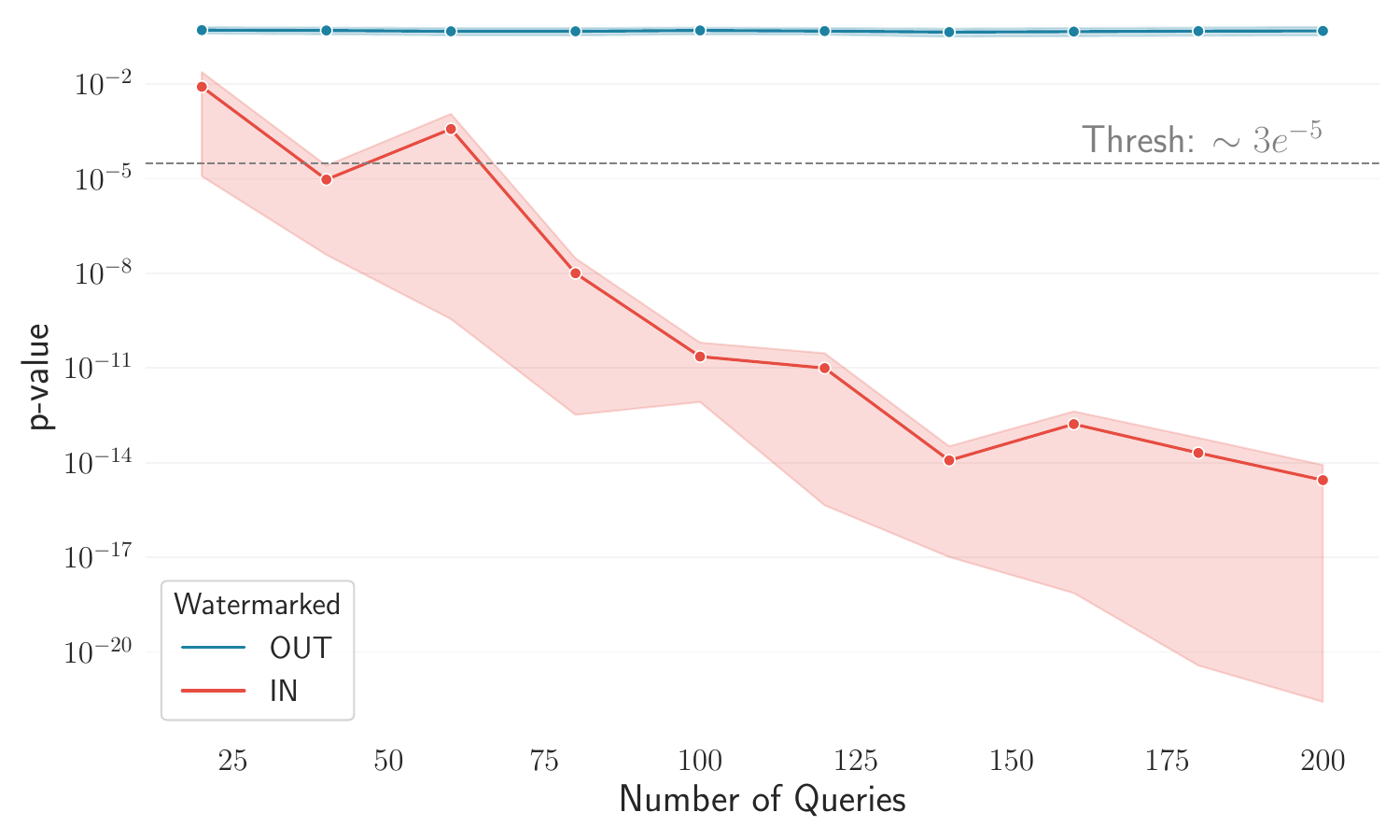}
    \vskip -0.5em
    \caption{Average p-values against \#queries in our imperfect retrieval experiment.}
    \label{fig:real_rag}
  \end{minipage}
\end{wrapfigure}

As above, we repeat all runs with $5$ seeds on both \easy{} and \hard{}, with $\llm=\llamaMid$ and both prompts.
We defer accuracy curves to~\cref{app:more_results:rag} but note that \ours{} achieves $100\%$  accuracy across all settings. 
In~\cref{fig:real_rag}, we show average p-values over the number of queries, further illustrating the reliability of \ours{}. 
We note that this simple RAG system was able to retrieve the targeted watermarked article in $93.6\%$ of all requests, almost perfectly reducing this to our idealized setting.
We further run all baselines with \defended{} in the \hard{} setting, and confirm that the (rare) retrieval errors do not make them more competitive, \eg because the distracting documents are now less relevant.
As in the corresponding part of~\cref{fig:main_exp}, \facts{} has 100\% false positives, while \sib{} and \ibm{} have 100\% true negatives.

\paragraph{Text quality}
\label{ssec:eval:quality}

\begin{wraptable}{r}{0.41\textwidth}
  \centering
  \vskip -1.3em
  \caption{Quality evaluation of watermarked documents. }
  \label{tab:quality}
  \vskip -0.5em 
  \resizebox{0.4\textwidth}{!}{%
      \begin{tabular}{lccc}
          \toprule
          & \textbf{Original} & \textbf{Paraphrased} & \textbf{$\Delta$} \\
          \midrule
          Judge & $0.903$ & $0.898$ & $5e^{-3}$ \\
          P-SP & $1.000$ & $0.933$ & $0.07$ \\
          \bottomrule
      \end{tabular}
  }
\end{wraptable}
Applying \ours{} requires the data owner to watermark $\ownerdata{}$, which we do by paraphrasing with a watermarked model.
While prior work shows that watermarks minimally impact text quality~\citep{markmy}, we re-affirm this by judging the quality of a randomly sampled subset of $100$ resulting documents. 
As shown in~\cref{tab:quality}, paraphrased documents maintain high quality both under \gptBig{} ratings (see~\cref{app:prompts:quality}) and a paraphrase quality metric P-SP~\citep{psp}.

We also evaluate the quality of the RAG system's responses (with \naive{} and \haiku{}) on $200$ samples.
The average response quality is $0.9475$ in \incase{} cases (watermarked context) and $0.9465$ in \outcase{} cases, confirming that watermarked documents do not negatively impact response quality.

\subsection{Ablations}
\label{ssec:eval:ablations}

\begin{figure}[t]
  \centering
  \begin{minipage}{0.49\linewidth}
    \includegraphics[width=\textwidth]{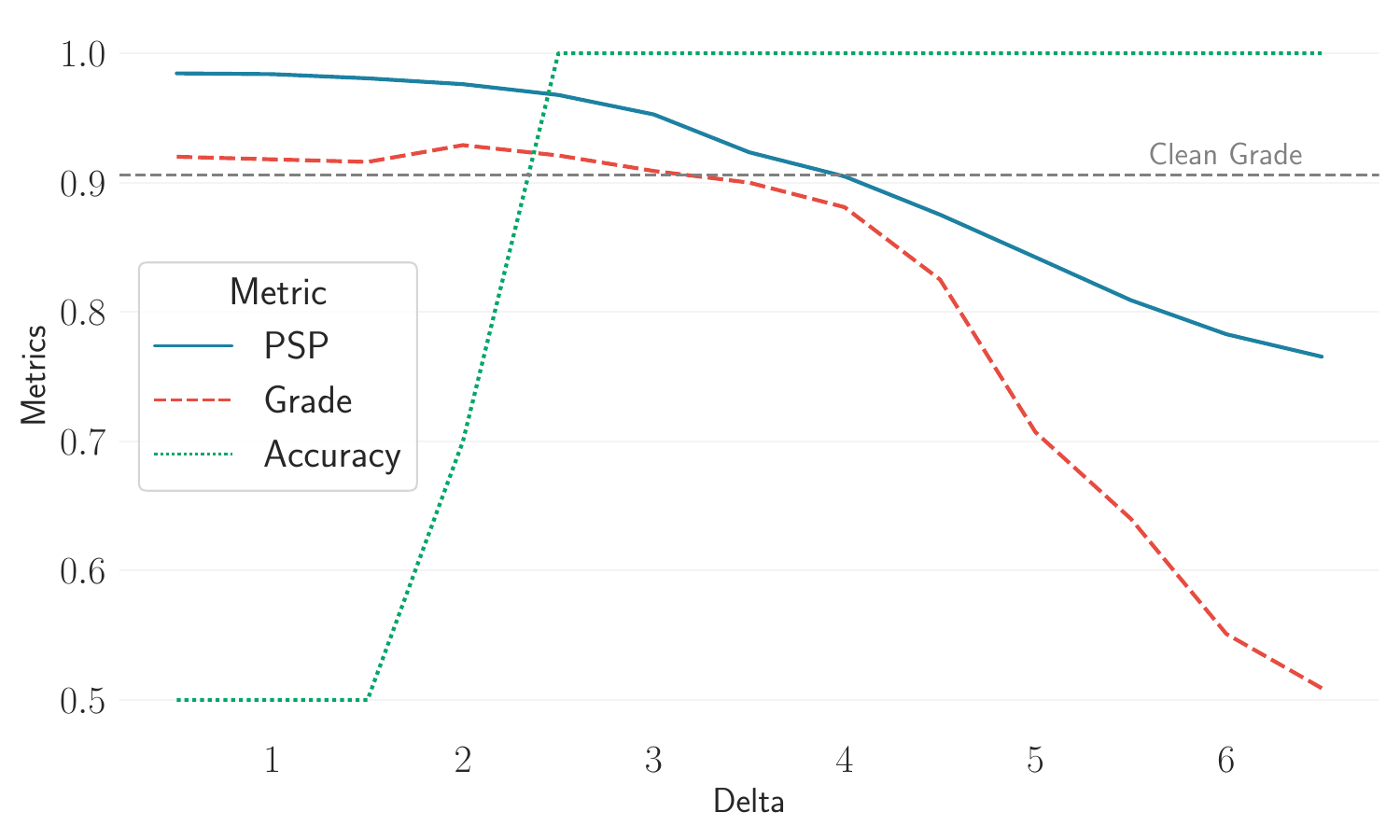}
  \end{minipage}
  \hfill
  \begin{minipage}{0.49\linewidth}
    \includegraphics[width=\textwidth]{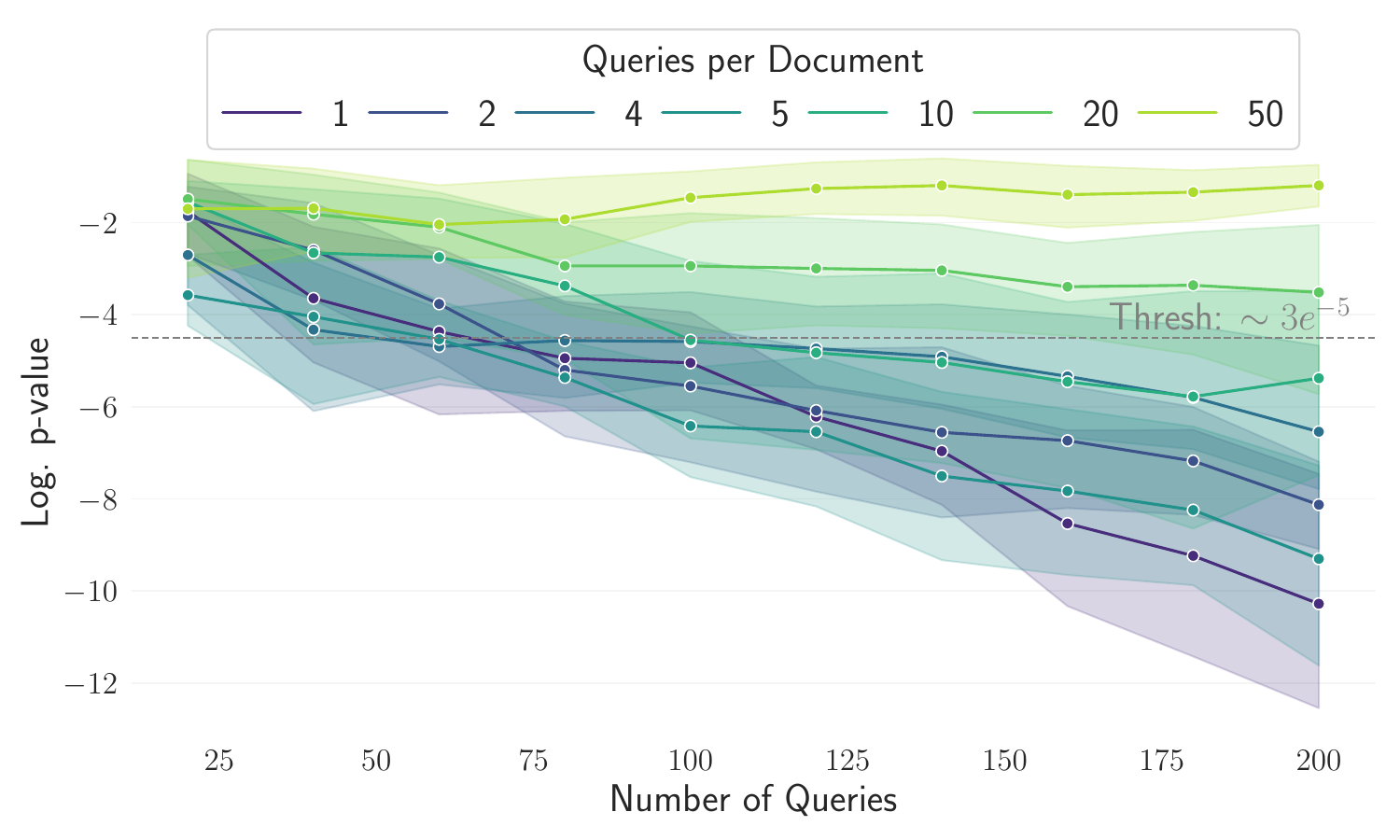}
  \end{minipage}
  \caption{Left: Accuracy and watermarked text quality for different $\delta$ values of \ours{} on \newdatah{}. Right: Accuracy over the number of queries per document for \ours{} on \newdatah{}. }
  \label{fig:watermark_params}
  \vskip -1em
\end{figure}

Lastly, we provide ablations for important hyperparameters of \ours{} and the \problem{} setup. 

\paragraph{Watermark parameters} 
\label{sssec:eval:ablation_parameters}

As introduced in \cref{sec:background}, red-green watermarks have $2$ key parameters: the strength $\delta$ and the context size $h$. 
We ablate over their impact on \ours{} using \llamaMid{} as $\llm$ with the \defended{} sysprompt. 
The results for $\delta$ are shown in \cref{fig:watermark_params} (Left).
We observe a \emph{sweet spot} of $\delta \in [2.5,4.5]$ with high accuracy (unlike $\delta < 2.5$) and the minimal impact on text quality (unlike $\delta > 4.5$).
Notably, experiments in~\cref{app:more_results:method} show that $\delta>4.5$ negatively affects both query efficiency and text utility.
This range of $\delta \in [2.5,4.5]$ directly aligns with recommendations in prior work~\citep{kgw2}, and supports our choice of $\delta=3.5$ for the main experiments. 

Regarding $h$, as noted in~\cref{sec:method} and prior work, larger values are expected to degrade robustness.
We compare our choice of $h=2$ with $h=4$, noticing a slight decrease in accuracy ($100\%$ on \easy{}, but $80\%$ on \hard{}), confirmed by our plots of query efficiency---full results are deferred to~\cref{app:more_results:method}.

\paragraph{\problem{} parameters}
The \problem{} setting has a wide range of parameters that we so far have not explored. 
First, we ablate over $k$, the number of retrieved documents---our results in~\cref{app:more_results:setting} show that \ours{} performs well independent of this parameter.
Also, in~\cref{app:more_results:setting}, we study the case where only a subset of $\ownerdata{}$ is included in $D$, relaxing the assumptions of \problem{}. While this naturally weakens the watermark signal, our results imply that \ours{} shows some robustness to this setting.

Finally, while we generally assume a single query per document $d_i$, we briefly study the potential of \ours{} to increase its query efficiency by extracting more value from each $d_i$. 
For this we introduce a parameter $qpd$ (\emph{queries per document}, $qpd=1$ in the main experiments), generating more questions for each $d_i$.
The main results for $qpd \in \{1,2,4,5,10,20,50\}$ are shown in~\cref{fig:watermark_params} (Right), where we plot the average p-value over the total number of \emph{queries}. 
As expected, for high $qpd$, additional queries for the same $d_i$ bring only marginal value---this is also confirmed in~\cref{app:more_results:qpd}, which shows how the effective number of tokens scored by the watermark detector increases slower for high $qpd$, as we encounter many repeated n-grams. 
However, for lower values such as $qpd=4$, reducing the required $|\ownerdata|$ $4$ times, we observe strong results, suggesting a promising avenue for future work.
\section{Related Work} \label{sec:related_work}

Closest related works to ours are \citet{sib} and \citet{ibm}, which propose membership inference (MI) for RAG that we adapt to \problem{} in~\cref{sec:setting}, and works that highlight the risk of MI in the related paradigm of in-context learning such as \citep{flocks}. 
Others study broader privacy and security aspects of RAG, such as poisoning to jailbreak the model or exfiltrate data~\citep{poisonedrag,badrag,phantom,goodbad}, concerns similar to \problem{}.

\paragraph{Passive MI/DI}
The problems of membership inference (MI)~\citep{mia,lira} or dataset inference (DI)~\citep{dia, diassl} have been long studied on training data, as opposed to RAG corpora as in this work.
Recent attempts to adapt these methods to LLMs~\citep{llmia1,llmia2,llmia,lldia} cite the challenge of rigorous evaluation, and primarily focus on graybox settings, citing the difficulty of inference attacks in the blackbox setting~\citep{labelonly}, which is what we consider in \problem{}.
Another perspective on the problem of tracing data usage in model training is given by recent works on LLM data contamination~\citep{dekonick,contam}, none of which consider RAG.
 
\paragraph{Proactive MI/DI/model protection}
Another approach to MI/DI is (as in this work) \emph{proactive}, e.g., by watermarking the data~\citep{entruth,domainwm}. 
However, only few works study LLMs, and only the ones cited above consider RAG.
\citet{activellm} focus on LLMs in a graybox setting, inserting random sequences or unicode substitutions to trace the data through training.
\citet{radioactivity} find that LLM watermarks propagate through fine-tuning, but \ownerdata{} has to make up 10\% of the fine-tuning corpus for blackbox detection.
\problem{} relaxes this requirement, as RAG is unaffected by $|D|$ and degrades the signal much less than fine-tuning.
Another orthogonal area is watermarking of \emph{models} against model stealing, often via backdoors~\citep{ip1,ip3}.

\paragraph{LLM Watermarking}
Finally, we note that many works study red-green LLM watermarks~\citep{kgw,kgw2,gptwm,semstamp}, but also other approaches such as sampling modification~\citep{stanford,orzamir}, model-based watermarking~\citep{private}, or watermarking in weights~\citep{learnability}.
We note that \ours{} could be combined with many of these, and leave this interesting direction to follow-up work. 
   
\section{Conclusion} \label{sec:conclusion}
We studied the problem of black-box RAG Dataset Inference (\problem{}), where the goal is to detect unauthorized usage of a dataset in a RAG system.
We formalized the problem, proposed a dataset and a set of baselines, and presented \ours{}, a method based on LLM watermarks, which provides rigorous statistical guarantees.
Our evaluation showed that \ours{} outperforms the baselines, achieving perfect accuracy and high query efficiency and robustness.
This establishes \ours{} as a practical tool that can be directly applied to protect the rights of data owners in current RAG systems.
We hope our work provides a valuable foundation for future work on \problem{}---interesting directions include combining \ours{} with other LLM watermarks, or designing watermarks specifically tailored to \problem{}.   

\clearpage %

\section*{Acknowledgements} \label{sec:acknowledgements}
The work has received funding from the Swiss State Secretariat for Education, Research and Innovation (SERI).

\section*{Ethics Statement} \label{sec:ethics}
We do not foresee any ethical concerns arising from our work.
Our work primarily provides dataset owners with the ability to prove unauthorized usage of their data in a RAG system, which is a valuable tool for protecting intellectual property, and fighting \emph{against} unethical practices of large model providers.
An undesirable side effect of such methods may come from false accusations---as our experiments in~\cref{sec:evaluation} demonstrate, our proposed method, \ours{}, strictly controls the rate of false positives as highly unlikely, and in all of our experiments we do not observe a single instance of a false positive.
While our work does result in a new dataset, \newdata{}, this dataset is entirely synthetic as it is based on fictional articles~\citep{repliqa}, and thus does not raise any privacy concerns.

\section*{Reproducibility Statement} \label{sec:reproducibility}
To foster reproducibility, we provide details of our experimental setup in~\cref{sec:evaluation} and \cref{app:expdetails}, and in~\cref{app:prompts} include all prompts used in our experiments.
We make our code and the \newdata{} dataset available at \url{https://github.com/eth-sri/ward} under an MIT license, and include configuration files for each experiment and a README file with instructions on how to reproduce our results.
\bibliography{references} 
\bibliographystyle{iclr2025_conference}
\vfill
\clearpage
   
\message{^^JLASTREFERENCESPAGE \thepage^^J}

\ifincludeappendixx
	\newpage
	\appendix
	\onecolumn 
	
\section{{Motivation: The Importance of RAG-DI}} \label{app:motivation}
{Extending on our discussion in~\cref{sec:introduction}, we elaborate on the motivation for establishing \problem{} as a problem setting, the prevalence of the problem of unauthorized data usage (studied in this work), the need for a solution like \ours{}, and its practical applicability.} 
{\paragraph{Protecting data from use in Generative AI}
The unauthorized usage of data in Generative AI (GenAI) systems, primarily for training, has become an especially pressing concern in recent years, as the practice of indiscriminate scraping of massive datasets from the web became more commonplace~\citep{genlaw_devil}.
This has led to a crisis in data transparency, where the provenance of data used to train models is often hard to track~\citep{longpre2023dataprovenanceinitiativelarge}, and commonly not reported by the model provider.
While the EU AI Act~\citep{aia} attempts to remedy this, by mandating that all data sources used for model training are published, this regulatory process is slow, and data owners must take active steps to protect themselves.}
{\paragraph{Opt-outs and legal uncertainties}
One tool that data owners can use to exercise agency over the use of their work comes in the form of opt-outs, where creators can choose to exclude their works from GenAI training.
Notably, the Article 4 of the EU's Copyright in Digital Single Market (CDSM) Directive~\citep{cdsm} recognizes the right of creators to exercise such an opt-out, by providing a machine-readable expression of the reservation of rights.
Yet, the practical implementation of this opt-out mechanism is still unclear, and an active discussion from both law~\citep{kluwerblog} and research~\citep{decorait} communities is ongoing. 
Many initiatives have established standards for expressing such reservations~\citep{optouts}, such as \texttt{ai.txt}~\citep{aitxt}, the TDM Reservation Protocol~\citep{tdmrep}, and C2PA~\citep{ctwopa}, a metadata standard for embedding content provenance information in media files which also includes a flag for opting out of GenAI training and/or inference.}

{While opt-outs are picking up traction, and some GenAI providers have pledged to honor them~\citep{spawning}, they are still far from being universally respected, and it is yet unclear how to practically enforce them (or audit compliance). 
More broadly, the legal landscape around GenAI data usage is complex, and both legal and research communities are closely following several prominent lawsuits.
Notable examples include NYT vs OpenAI/Microsoft~\citep{nyt} or a group of artists vs Stability/MidJourney/DeviantArt/RunwayML~\citep{imagecopyright}, which illustrate that the legal system is naturally still catching up to the rapid development of GenAI. 
The key challenges are in untangling the interplay between licenses, traditional copyright, opt-outs, the ideas such as fair use or market harm, and technical concepts such as LLM memorization.
The most relevant for our discussion is the case Kneschke vs LAION~\citep{kneschkelion}, where the ruling has extensively engaged with the meaning of opt-outs and the extent of their enforceability.}
{\paragraph{Proving unauthorized usage}
In this landscape of legal uncertainty, the ability to \emph{provably detect} unauthorized usage of data in GenAI systems is crucial for data owners, as it may provide a way to enforce their rights in court. 
While this is relatively simple when model providers make their datasets public~\citep{haveibeentrained}, the problem becomes significantly harder for the common case of undisclosed training data, leading to works on proactive membership/dataset inference such as~\citet{WeiWJ24,radioactivity} and others we cite in~\cref{sec:related_work}.
As noted in~\cref{sec:introduction}, a particular flavor of this problem comes in the context of LLMs, where RAG systems have become a common way to integrate new knowledge into the model without costly fine-tuning.}

\paragraph{\textsc{Ward}}
{This is the problem statement studied in this work and tackled by \ours{}, with our requirements (\cref{sec:setting}) directly inspired by the legal and practical challenges data owners face in this context.
Namely, \emph{Guarantees} are crucial to be able to use the results as evidence, and \emph{Robustness} is important to ensure that the method is not easily circumvented, unintentionally or by attempts to defend against it.
Finally, \emph{Monotonocity} and the corresponding \emph{Efficiency} are needed for the method to be practically feasible. 
While the data owner naturally must query every system they suspect is using their data in the RAG corpus, each such use is very cheap (below $\$2$ per LLM, see~\cref{app:efficiency}).
Further, we argue that the ecosystem of relevant LLM providers is relatively small. In particular, not many providers both (i) have resources for indiscriminate large-scale scraping of data and (ii) are sufficiently popular to create market harm based on the unauthorized usage of that data.
Thus, the cost of \ours{} is practical.} 

\section{Additional Experimental Details}
We expand on the details of our experimental setup provided in the main paper, by providing a detailed explanation of how we sample \newdata{} to create our evaluation settings (\cref{app:expdetails}), and how we simulate perfect retrieval in our experiments (\cref{app:retrieval}).

\subsection{Sampling \newdata{} for Evaluation} \label{app:expdetails}
Recall that each \newdata{} \emph{group} contains $4$ articles, one for each of the LLM authors we consider.
For brevity, we will use $\mathcal{A}_i$ for $i \in \{1,2,3,4\}$ to denote the $i$-th author.
For the \easy{} setting, we sample four {subsets} of {distinct} groups, where the sizes of the subsets are respectively $(200, 300, 300, 200)$.
Then for each subset $i \in \{1,2,3\}$, we only take articles from $\mathcal{A}_i$, and include all of them in the RAG corpus $D$.
Out of those, articles from subset $1$ of author $\mathcal{A}_1$ are taken as $\ownerdata{}$ in the \incase{} case, i.e., these are potentially modified by the service provider before inserting them into the RAG corpus.
Similarly, the articles from subset $4$ of author $\mathcal{A}_4$ are reserved as $\ownerdata{}$ in the \outcase{} case.
This setup ensures no fact redundancy in the RAG corpus.

In contrast, to create the \hard{} setting with fact redundancy, we start by sampling $1000$ distinct groups.
The RAG corpus $D$ is then built by including all articles from those groups that were written by $\mathcal{A}_1$, $\mathcal{A}_2$, and $\mathcal{A}_3$.
A randomly subsampled set of $200$ of those articles, that were written by $\mathcal{A}_1$, is taken as $\ownerdata{}$ in the \incase{} case.
Similarly, randomly sampling $200$ of the $1000$ above groups, and taking documents from $\mathcal{A}_4$ out of each group is used as $\ownerdata{}$ in the \outcase{} case.
Due to practical limitations, before running all of our experiments we managed to generate only the first $1000$ out of $3391$ groups in \newdata{}, thus our sampling is done w.r.t. these $1000$ groups.

\subsection{{Simulating Perfect Retrieval}} \label{app:retrieval}
{We provide a detailed description of the perfect retrieval mechanism used in our main experiments, as introduced in~\cref{sec:evaluation}, and further compared to a real retrieval mechanism in~\cref{ssec:eval:add_considerations}, where we have demonstrated that it almost perfectly matches the idealized case described in this section and shown that \ours{} performs equally well in both cases.}
{Assume the data owner is querying $\ragsys$ in order to test for presence of a document $d$. We construct the following sequence of documents:
\begin{itemize}
    \item First, the document $d$ itself if it is present in the RAG corpus.
    \item Next, a random shuffle of all documents $d'$ from the same $\newdata{}$ group that are present in the RAG corpus, i.e., the documents produced by other \emph{authors} based on the same \emph{source}.
    \item Finally, a random shuffle of all other documents from the RAG corpus.
\end{itemize}
Given the parameter $k$ of desired \emph{shots} (ablated in~\cref{fig:shots}), the perfect retrieval returns the first $k$ documents from this sequence. 
Note that for $\ours$ and $k>1$ this always results in additional documents in the context beyond the watermarked one, where such distractions often come from both the same source, as well as different sources.
We further illustrate the behavior of perfect retrieval with two examples.}
{\paragraph{First example}
The RAG provider populates the corpus $D$ according to the \easy{} setting, and uses $k=3$. 
The user applies \ours{} to query $\ragsys{}$ with a question related to the (watermarked) document $\#123a$, i.e., from \newdata{} group $\#123$, written by the author $a$.
Assume $\#123a \in D$.
The retrieved documents are $\#123a$, $\#374b$, $\#266a$.
Following the above steps, we include $\#123a$ as it is present in the corpus, and two documents chosen at random, as in the \easy{} setting the corpus includes no other document from group $123$.}
{\paragraph{Second example}
Now assume the \hard{} setting and $k=5$ shots.
The user queries $\ragsys{}$ with a question related to $\#123d$.
Assume $\#123d \notin D$.
The retrieved documents are $\#123a, \#123b, \#123c, \#597c, \#442b$.
Per above steps, as the target document $\#123d$ is not in the corpus, we include all other documents from group $\#123$ and $2$ documents chosen at random. }   
\section{Additional Experimental Results} 
\label{app:moreresults}
In this section, we first present additional results on our \problem{} ablations, omitted from~\cref{ssec:eval:ablations} (\cref{app:more_results:setting}), 
Similarly, in~\cref{app:more_results:method}, we extend our \ours{} ablations from~\cref{ssec:eval:ablations}.
In \cref{app:more_results:rag} we provide extended results of our retrieval study from~\cref{ssec:eval:add_considerations}, and in \cref{app:more_results:qpd} extend our results on the number of queries per document from~\cref{fig:watermark_params} (Right).
Finally, in \cref{app:more_results:memfree} we provide results on an additional defense (\textsc{MemFree}, summarized in~\cref{ssec:eval:desiderata}), and in \cref{app:more_results:baselines} we provide more insights into the performance of \problem{} baselines. 

\subsection{\problem{} Ablations}
\label{app:more_results:setting}

\begin{figure}[t]
  \centering
  \begin{minipage}{0.48\linewidth}
    \includegraphics[width=\textwidth]{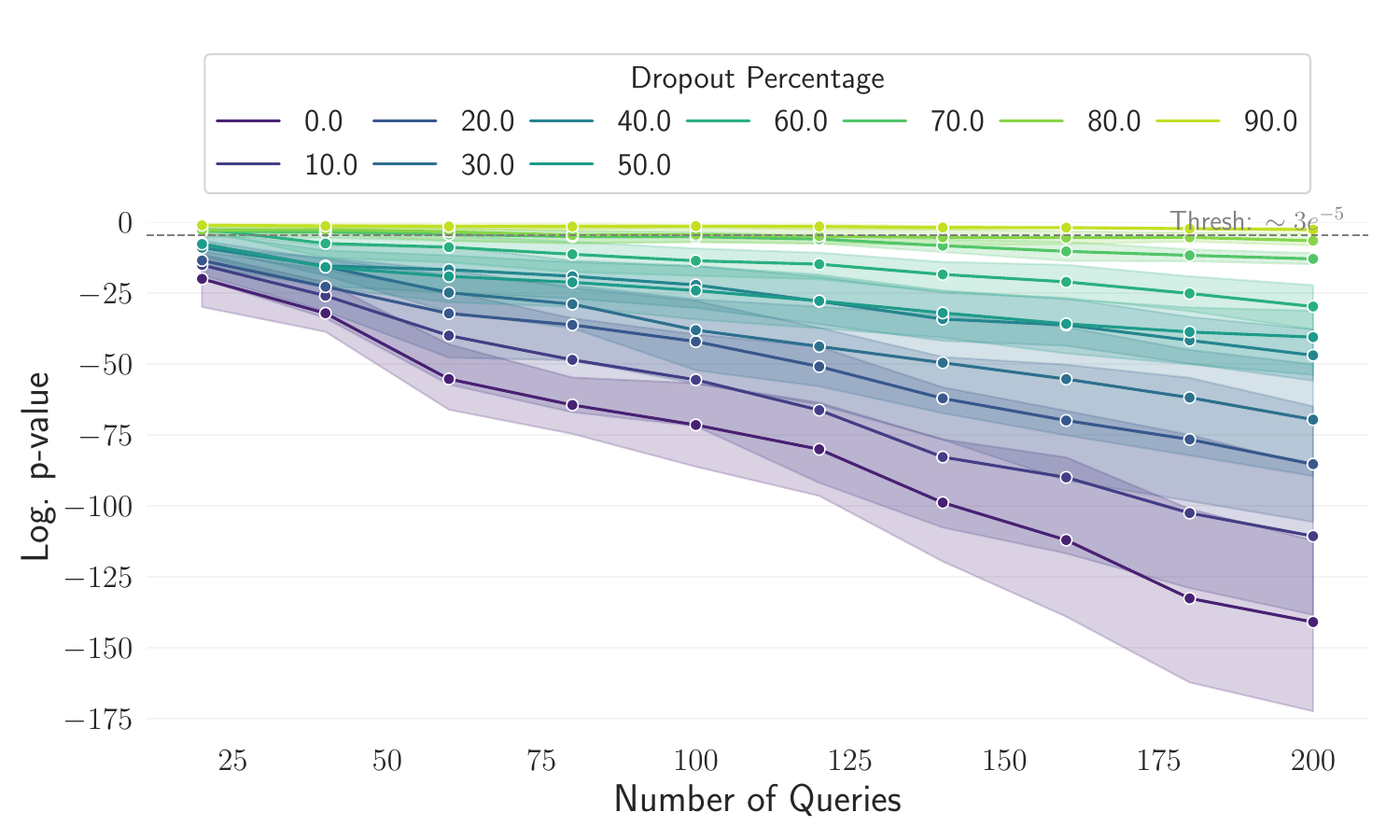}
  \end{minipage}
  \hfill
  \begin{minipage}{0.48\linewidth}
    \includegraphics[width=\textwidth]{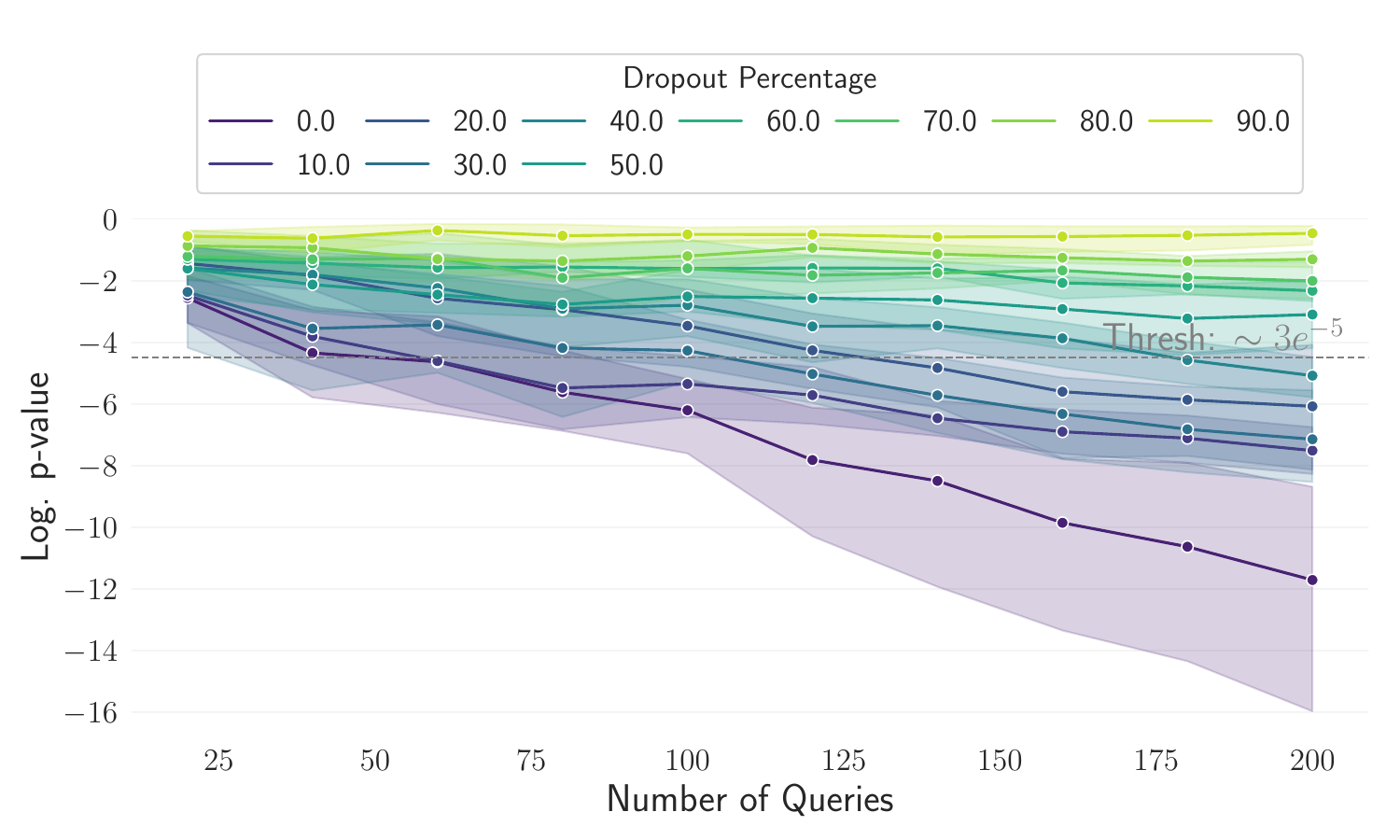}
  \end{minipage}
  \caption{Effect of the percentage of documents of $\ownerdata$ \textbf{not} contained in $D$ on the p-value for \llamaMid{} using \naive{} (left) and \gptSmall{} using \defended{} (right) on \newdatah{}. While in both cases we see how \ours{} is robust to an increase in the percentage of documents contained in $D$, the \defended{} setting is more sensitive, showing initial false positives at $20\%$.}
  \label{fig:abl_dropout}
\end{figure}

\begin{wrapfigure}[17]{R}{0.47\textwidth} 
  \centering
  \vskip -1.4em
  \begin{minipage}{\linewidth}
    \includegraphics[width=\textwidth]{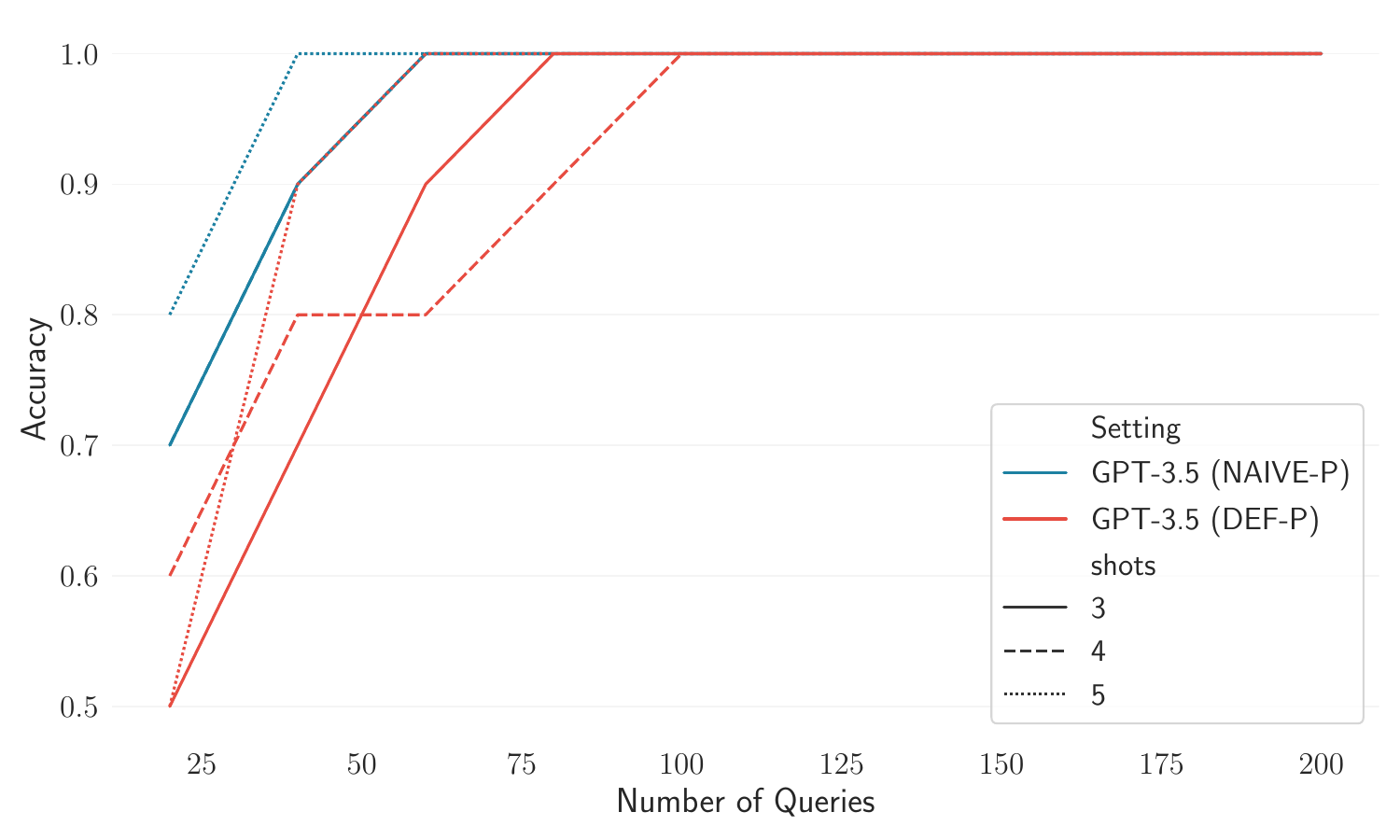}
    \caption{Accuracy over the number of queries when varying the number of retrieved documents in RAG (shots, $k$) for \ours{} on \newdatah{}. We observe, on \gptSmall{}, that \ours{} scales well with the number of documents for both the \naive{} and \defended{} system prompts.}
    \label{fig:shots}
  \end{minipage}
\end{wrapfigure}

As summarized in~\cref{ssec:eval:ablations}, we ablate over key parameters of \problem{}: The number of documents (shots) put into the context of model ($k$) and the fraction of documents of $\ownerdata$ contained in $D$.

\paragraph{Number of shots}
For $k$ we present our results for \ours{} in \cref{fig:shots}, using $k \in \{3,4,5\}$ for our \naive{} and \defended{} settings on \newdatah{} using \gptSmall{}. 
Across all values of $k$, \ours{} shows favorable scaling of the number of queries made to the system, reaching $100\%$ accuracy at $100$ queries at the latest. 
This follows intuition, as the scaling of \ours{} is primarily influenced by the capabilities of the underlying LLM to select the correct information from a set of retrieved documents. 
The constant improvement in model capabilities, therefore, naturally positively impacts the scaling of \ours{}  w.r.t. context size. 
In all cases, \ours{} reaches $100\%$ accuracy at a rate matching that of our main experiments.
While the above values of $k$ are generally recommended~\citep{flashrag}, and used in prior RAG MIA work~\citep{sib,ibm}, we additionally experiment with a much larger $k=10$, switching to \haiku{} (as this experiment exceeds \gptSmall{}'s context size).
In all cases, \ours{} reaches $100\%$ accuracy at a rate matching that of our main experiments.

\paragraph{Percentage of documents in the corpus}
We further ablate over the percentage of documents of $\ownerdata$ contained in $D$, which our main experiments assume to be $100$.
For this extension of the \problem{} setting, we assume that while the data owner wants to check whether $\ownerdata \subseteq D$, in reality only a strict subset $\ownerdata' \subset \ownerdata$ with size $|\ownerdata'|= (1-\omega) \cdot |\ownerdata|$ is contained in $D$. 
Naturally, as the data owner in \ours{} incorporates all queries for all documents, the resulting queries related to documents not from $\ownerdata'$ will increase the p-value. 
We experimentally test this in two settings: once with \llamaMid{} and \naive{}, and once with \gptSmall{} and \defended{}, presenting the mean $log_{10}$ p-value of the number of queries for both settings in \cref{fig:abl_dropout}. 
Across both cases, we observe that (as expected) a higher value of $\omega$ results in a higher computed p-value and hence a weaker test. \
At the same time, we can see that, especially in easier settings, \ours{} can endure a significant drop (only $20\%$ of $\ownerdata$ being contained in $D$) while still providing accurate results. 
In the more challenging setting, we observe the first false positives at $\omega = 0.2$, which both highlight the robustness of $\ours{}$  and provide an interesting avenue for future work.

\subsection{\ours{} Ablations}
\label{app:more_results:method}

To supplement our results in~\cref{ssec:eval:ablations}, we present full results of our ablation experiments of key watermarking parameters in \ours{}: the watermarking strength $\delta$ and the context size $h$.

\begin{figure}[t]
  \centering
  \begin{minipage}{0.49\linewidth}
    \includegraphics[width=\textwidth]{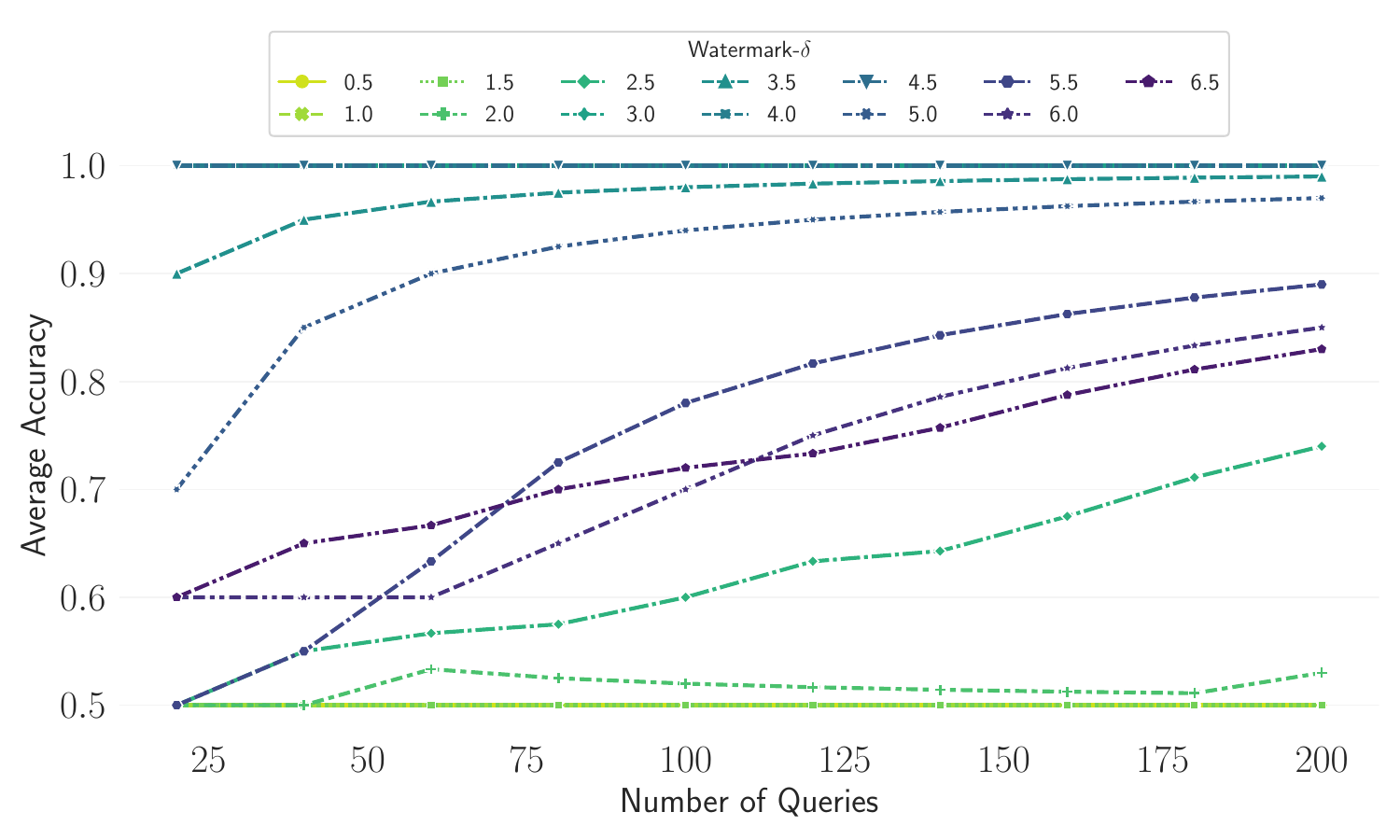}
    \caption{Average \emph{cumulative} accuracy across $\delta$s for \ours{} with \naive{} on \newdatah{}, where the model is \llamaMid. We observe an optimum at $\delta \approx 4$.}
    \label{fig:app_delta_ablation}
  \end{minipage}
  \hfill
  \begin{minipage}{0.49\linewidth}
    \includegraphics[width=\textwidth]{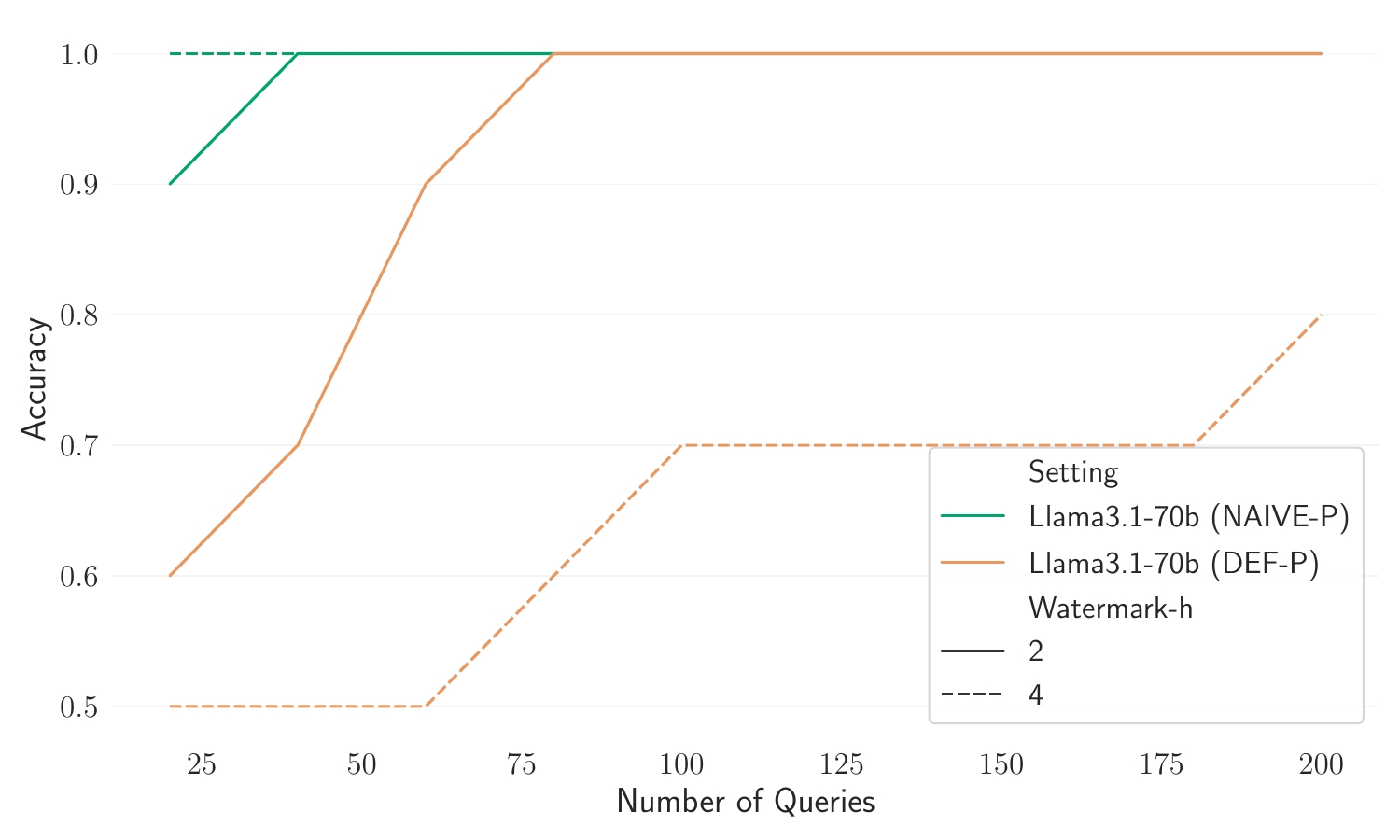}
    \caption{Accuracy for $h\in\{2,4\}$ using \ours{} on \llamaMid{} on \newdatah{}. We observe that \ours{} requires more samples to reliably detect unauthorized usage at higher $h$.}
    \label{fig:h_scaling}
  \end{minipage}
\end{figure}

\begin{wrapfigure}[26]{R}{0.46\linewidth}
  \vspace{-0.35in}
  \centering 
  \begin{minipage}{\linewidth}
    \includegraphics[width=\linewidth]{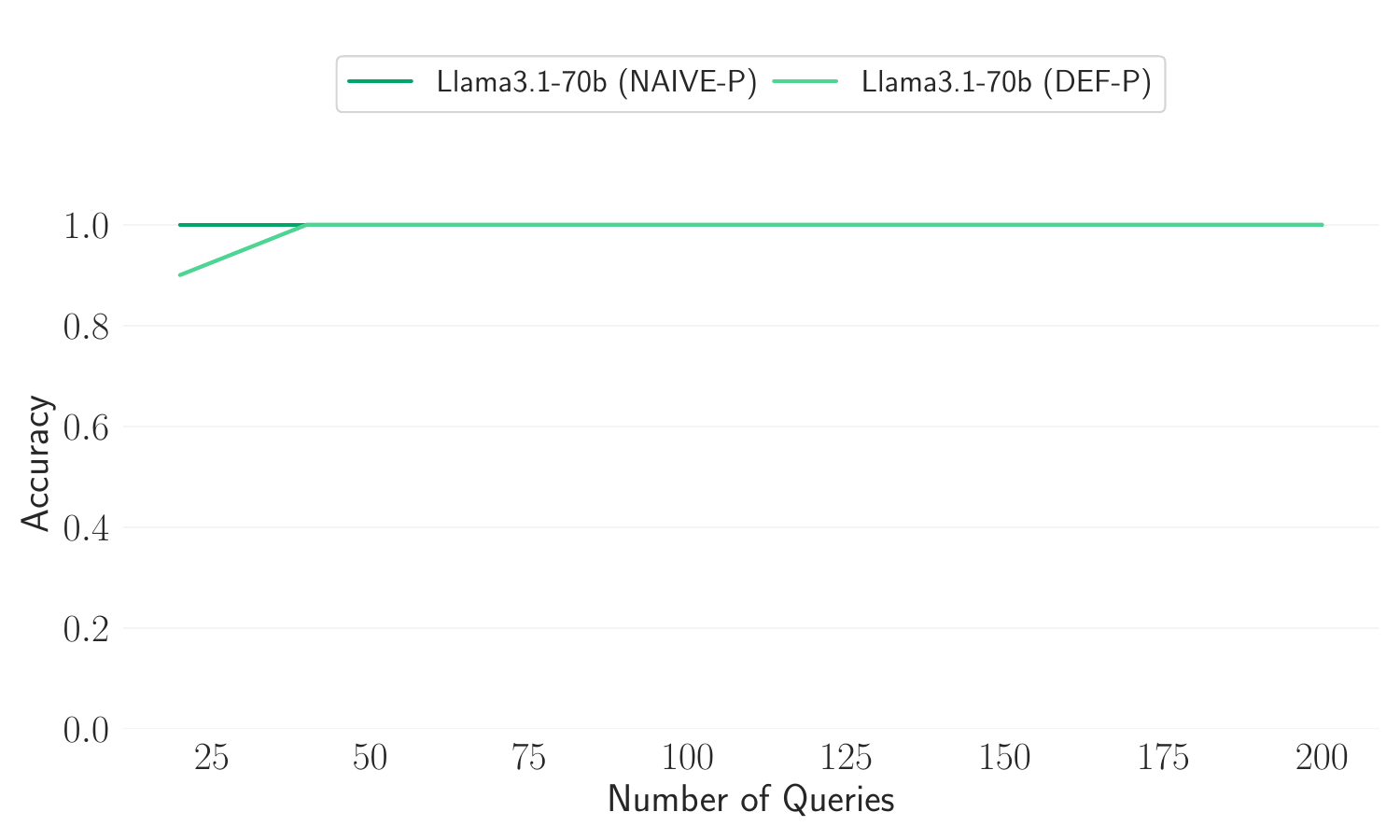}
  \end{minipage}
  \begin{minipage}{\linewidth}
    \includegraphics[width=\linewidth]{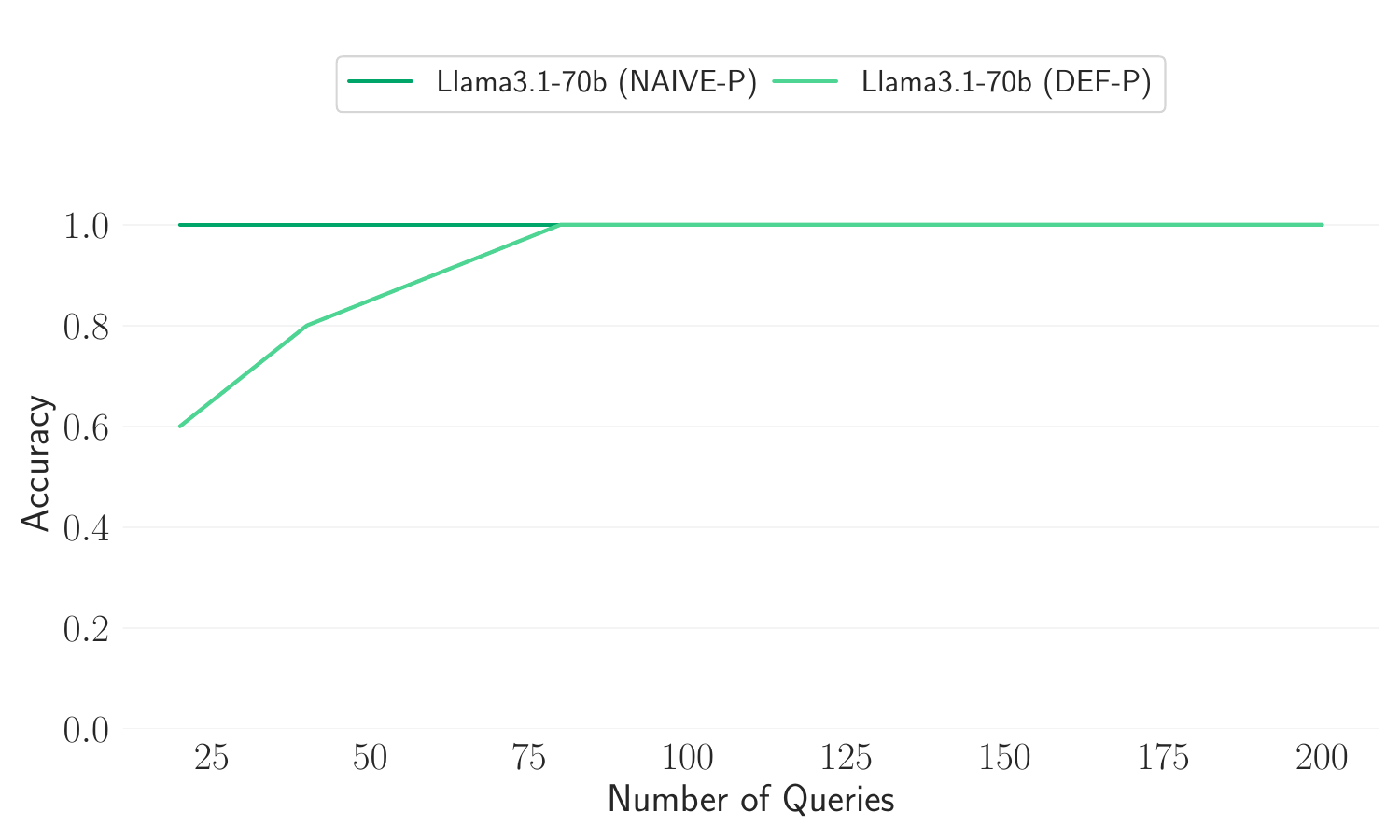}
  \end{minipage}
  \caption{Accuracy of \ours{} on \newdatae{} (Top) and \newdatah{} (Bottom) using a full RAG system with \llamaMid{}. As in our perfect retriever setting, we observe that \ours{} requires only a small number of queries to achieve 100\% accuracy in both settings.}
  \label{fig:real_rag_abl}
\end{wrapfigure}

\paragraph{Watermark strength}
We ablate $\delta \in [0.5, 6.5]$ in steps of $0.5$ using  \llamaMid{} with \defended{} on \newdatah{}. 
In particular, in \cref{fig:app_delta_ablation}, we complement our quality-accuracy plot from \cref{sssec:eval:ablation_parameters}, by displaying the \emph{cumulative} average accuracy at each number of queries. 
Concretely, we, at point $x$, present the average accuracy of all previous $x' \leq x$. 
This highlights two key findings.
First, while many higher $\delta$s achieve 100\% accuracy in our plot in \cref{fig:watermark_params} (as number of queries is $200$), they actually achieve worse results for a lower number of queries.
This can be explained by the fact that higher $\delta$s lead to worse text quality, which in turn impacts the text quality in the final LLM responses, and thus reduces the amount of watermark signal that is transferred. 
Second, these results narrow the optimal range of $\delta$ for our setting to $[3.5,4.5]$, on which we consistently achieve the best results.

\paragraph{Context size}
Further, we ablate over the watermark context size $h$, presenting additional results on $h=4$ in \cref{fig:h_scaling} (\llamaMid{} on \naive{}). 
Notably, we find, in line with prior work on watermarks showing that increases in $h$ produce less robust watermarks, that \ours{} requires more samples to reliably detect higher $h=4$ (compared to $h=2$). 
While on \easy{}, this has only a minor impact, we see a stronger initial drop on \hard{}. 
At the same time, \ours{} shows a strong monotonic increase even on \hard{}, highlighting its robustness.

\subsection{End-to-End RAG}
\label{app:more_results:rag}

Next, we complement our results on the full RAG implementation from \cref{ssec:eval:add_considerations} in \cref{fig:real_rag_abl}. 
For this, we show the full-accuracy curves on both \newdatae{} and \newdatah{} using an end-to-end RAG system as described in~\cref{ssec:eval:add_considerations}.
As expected, we observe that \ours{} requires only a few of queries ($40$ for \newdatae{} and $80$ for \newdatah{}) to achieve 100\% accuracy. 
This not only confirms our optimal retriever assumption in \cref{sec:evaluation} but also highlights how \ours{} is practical in real-world settings.

\begin{figure}[t]
  \centering
  \begin{minipage}{0.49\linewidth}
    \includegraphics[width=\textwidth]{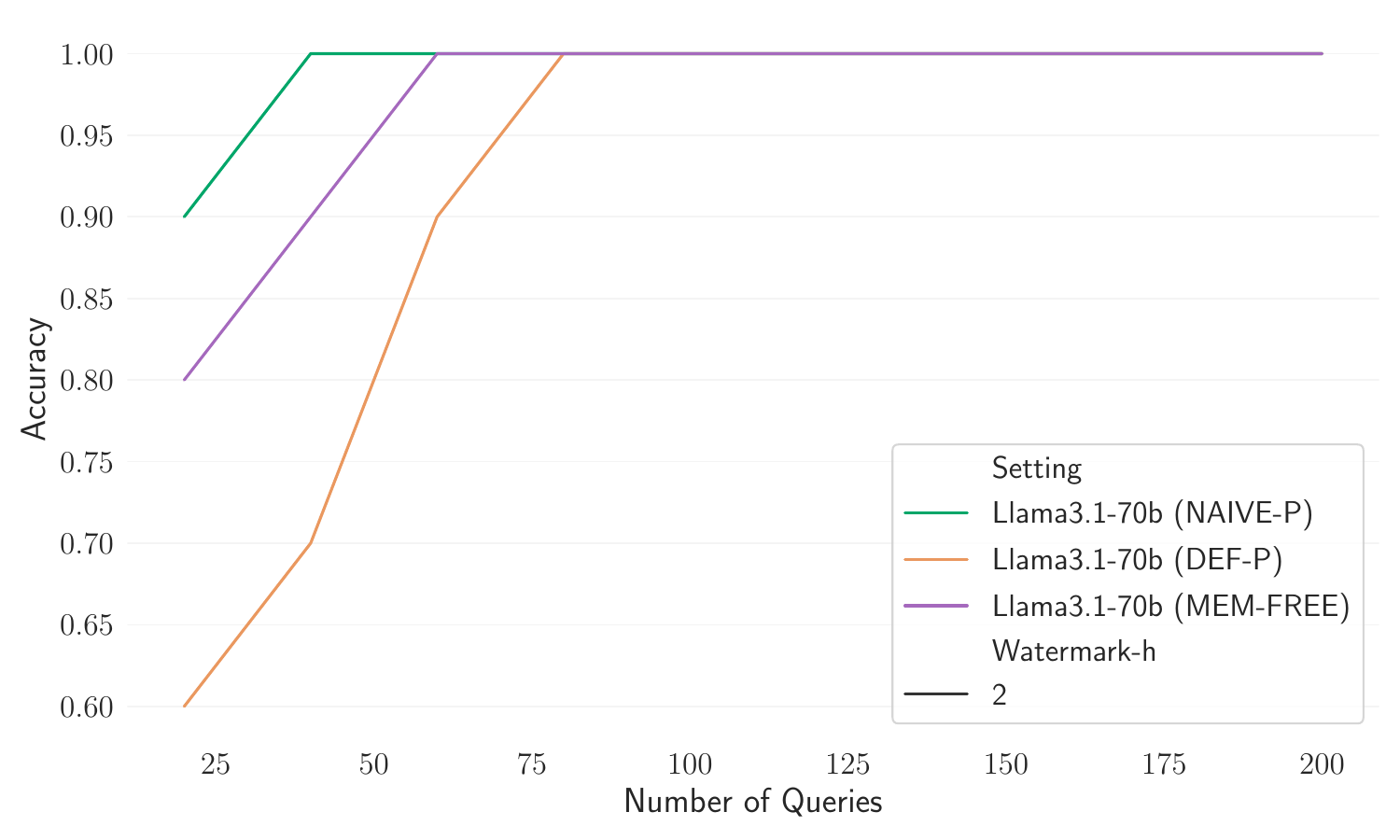}
  \end{minipage}
  \hfill
  \begin{minipage}{0.49\linewidth}
    \includegraphics[width=\textwidth]{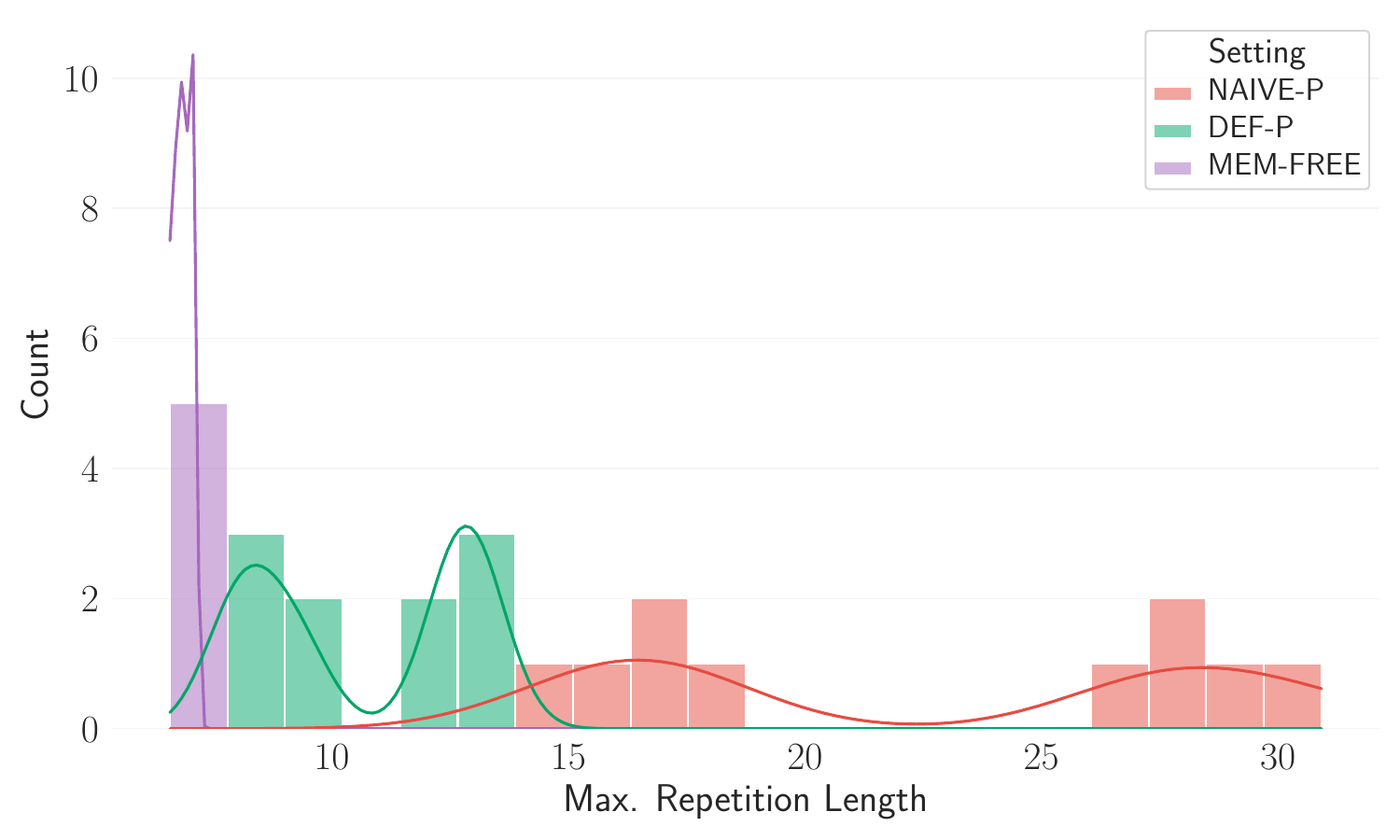}
  \end{minipage}
  \caption{Experiments with \textsc{MemFree}. \textbf{Left}: Accuracy of \ours{} on \newdatah{} and \llamaMid{} under the \textsc{MemFree} defense. Despite this more radical defense we observe that \ours{} requires only $60$ queries to achieve $100\%$ accuracy. \textbf{Right}: Maximum n-gram overlap between the retrieved documents and the generated responses for \gptSmall{} on \newdatah{} under the \textsc{MemFree} defense.}
  \label{fig:memfree}
\end{figure}

\begin{wrapfigure}{R}{0.5\textwidth}
  \centering 
  \vskip -1.6em
  \begin{minipage}{\linewidth}
      \includegraphics[width=\textwidth]{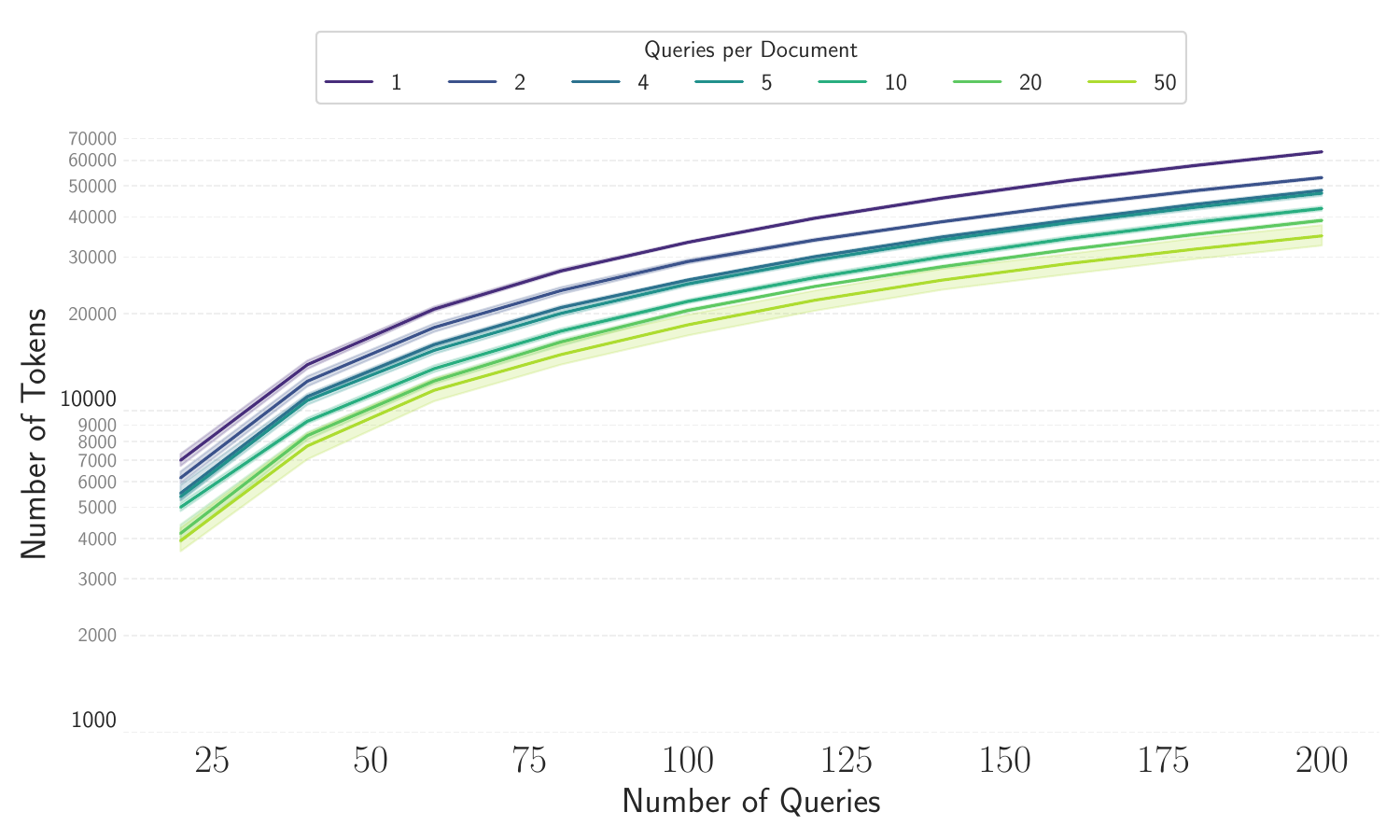}
      \caption{Number of usable tokens over \#queries for varying number of \emph{queries per document} in the \incase{} case, using \gptSmall{} and \defended{} in \newdatah{}.}
      \label{fig:tokens_per_qpd}
  \end{minipage}
\end{wrapfigure}

\subsection{Number of queries per document}
\label{app:more_results:qpd}

As presented in \cref{ssec:eval:ablations}, a data owner could phrase multiple queries per document in $\ownerdata$ to improve efficiency.
While we presented generally diminishing returns under a fixed query budget in \cref{fig:watermark_params}, we reaffirm this here by showing the actual number of tokens present in the resulting outputs that are actually \emph{scored} by the watermark detector after deduplication.
For this, we assume the \incase{} case using \gptSmall{} and \defended{}. 
We show the corresponding plot in \cref{fig:tokens_per_qpd} (log scale), and note that while we observe consistently linear scaling across all numbers of queries per document, reusing the same document too many times can lead to a decrease in the number of usable tokens (around $2$x for the highest setting of $50$ queries per document).
This is in line with our intuition, as obtaining unique watermarked n-grams is limited by the watermark of the original document that is present in the RAG corpus, and those tokens eventually get exhausted.
Still, as our results above have shown, a number of queries per document can likely be increased to at least $4$ to lead to a multiplicative increase in efficiency in terms of needed dataset size.

\subsection{\textsc{MemFree} Defense}
\label{app:more_results:memfree}

Inspired by~\citep{memfree}, we evaluate \ours{} on an additional defense, \textsc{MemFree}, which prevents the RAG system from producing any output that has a certain n-gram overlap with any of the retrieved documents.
To this end we adapt the procedure of \citep{memfree} to our setting, setting the maximum n-gram overlap to $10$ as in their work.
We can directly observe the effectiveness of this defense in reducing n-gram overlap in \cref{fig:memfree} (Right). 
At the same time, we find that \ours{} is very robust against such blunt defenses.
In particular, as we show in \cref{fig:memfree} (Left) on \newdatah{}, \ours{} requires only $60$ queries in order to achieve $100\%$ accuracy, which is only slightly more than in the undefended case.
We draw two conclusions from this: (1) \ours{} is surprisingly robust even in the face of stronger defenses, and (2) the search for stronger defenses or (then inversely) stronger dataset inference methods is a promising field for future research.

\subsection{Further Study of Baselines}
\label{app:more_results:baselines}

All our baselines introduced in \cref{ssec:setting:baselines} are specifically tuned on a training dataset to obtain the (empirically) optimal decision boundaries. 
However, as we find upon closer inspection, especially in harder settings, there is no \emph{single optimal threshold} to choose. 
We substantiate this in \cref{fig:threshes}, where on \sib{}, we show two regions w.r.t. the threshold: \textit{nontrivial} (gray; $>50\%$ accuracy for all settings/models, note that obtaining $50\%$ is trivial by simply always making the same decision, that $\ownerdata$ is in $D$ or otherwise), and \textit{optimal} (green; $100\%$ for all settings/models).

Notably, while we find that we can find an optimal threshold on \newdatae{} around $0.4$, finding a single such threshold for \newdatah{} is not possible. We observe this for all baseline methods in \newdatah{}, which helps explain why they fail to achieve a consistently high accuracy across our main experiments (where it is sensible to choose only a single threshold).

\begin{figure}[t]
  \centering
  \begin{minipage}{0.49\linewidth}
    \includegraphics[width=\textwidth]{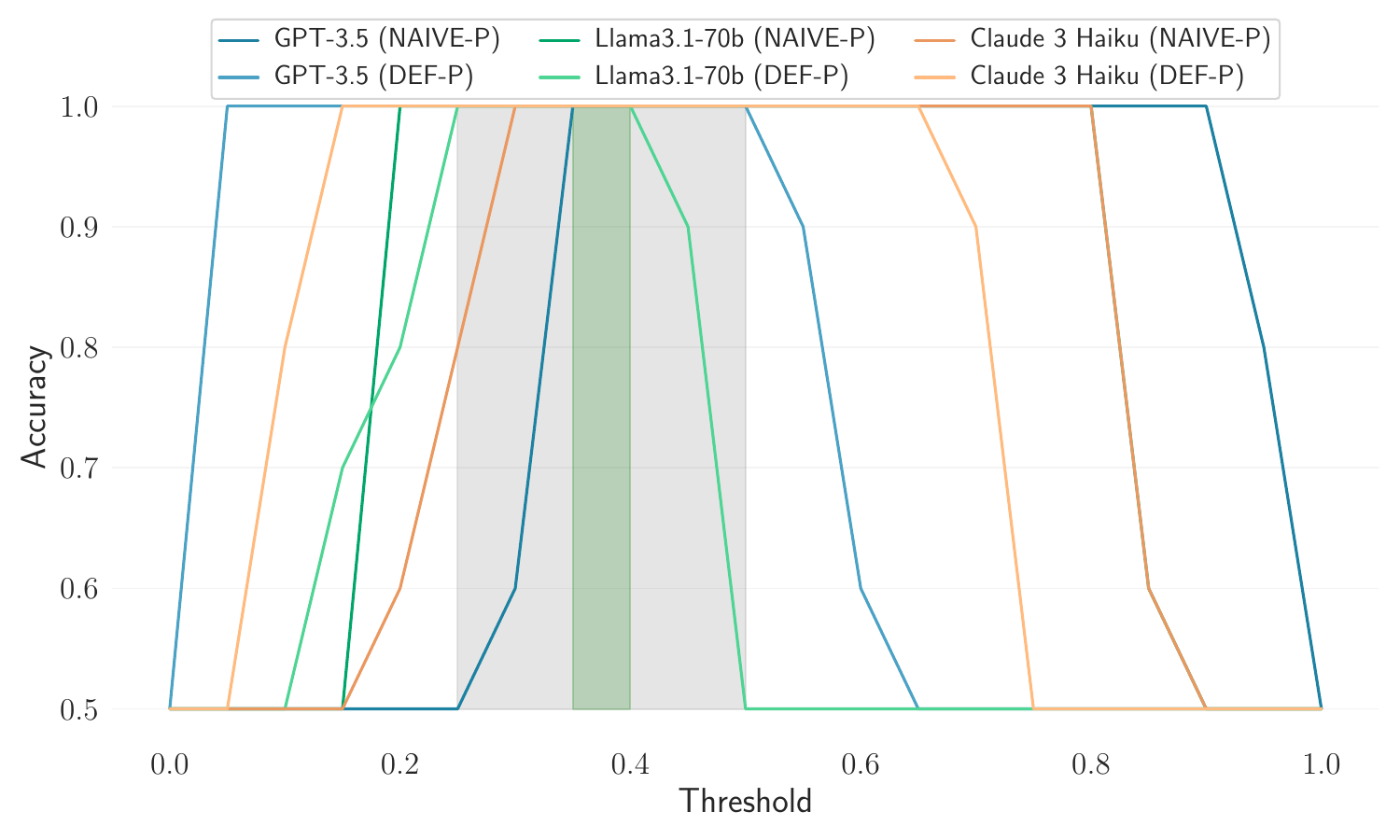}
  \end{minipage}
  \hfill
  \begin{minipage}{0.49\linewidth}
    \includegraphics[width=\textwidth]{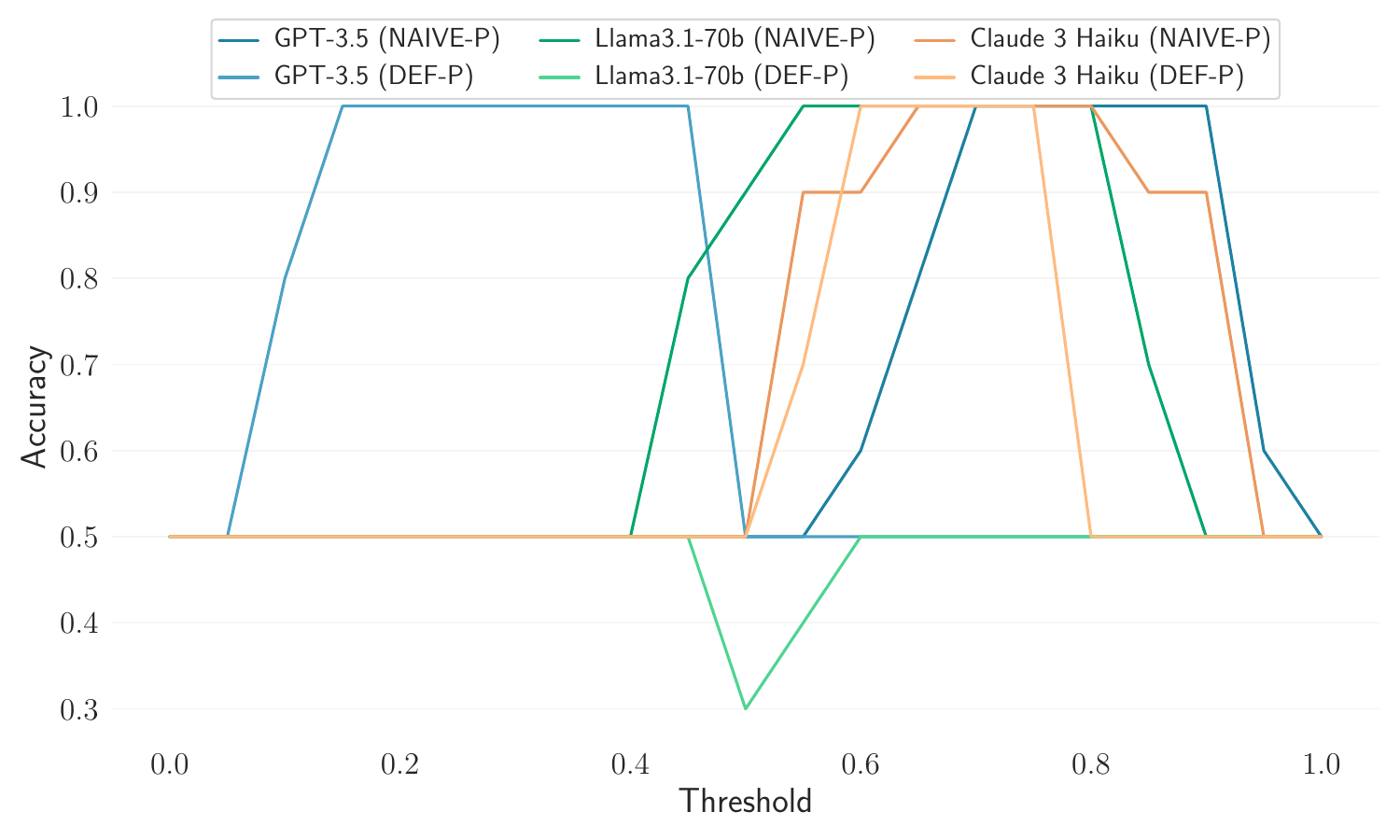}
  \end{minipage}
  \caption{Accuracy of \sib{} on \newdatae{} (left) and \newdatah{} (right) using all settings. We show the nontrivial (gray) and optimal (green) thresholds for \sib{} on \newdatae{} and \newdatah{}. We find that while a single threshold is viable for \newdatae{}, it is not for \newdatah{}.}
  \label{fig:threshes}
\end{figure} 

\section{Efficiency of \ours{}} \label{app:efficiency}
While not a separate point in the list of design requirements we introduce in~\cref{sec:setting}, we have extensively studied \emph{Efficiency} through the requirement of \emph{Monotonocity}, requiring that the accuracy of \ours{} increases \emph{fast} with more queries.
Namely, in~\cref{fig:scaling} we have discussed the efficiency of \ours{} on an intuitive level, and have confirmed it experimentally in~\cref{ssec:eval:desiderata}, where most of our experiments explore scaling with the number of queries, showing that at most $100$ queries to the RAG system are sufficient for a confident decision across all settings, models and defenses. 

{While we have focused on the {number of queries} as the primary metric, we can translate this into a more practical measure of API costs. 
We empirically observe that the queries $q_i$ sent by \ours{} to the RAG system $\ragsys{}$ are $30$ to $80$ tokens long, and the RAG system responses $r_i = \ragsys(q_i)$ are $400$ to $500$ tokens long.
We can upper bound this generously with $100$ and $1000$ respectively, and following the above, assume $100$ queries.
Given the November 2024 costs of GPT-3.5 ($\$3$ per 1M input tokens, $\$6$ per 1M output tokens), this results in a cost of only $\$0.63$ for the whole process of \problem{}. 
Repeating the calculation using the API costs of other popular closed-source models, the total cost of \problem{} with \ours{} is never above $\$2$.}

{An important consideration is also how the efficiency of \ours{} is impacted by the size of the RAG system's corpus $D$ and the data owner's dataset $\ownerdata{}$.
First, as we note in~\cref{sec:related_work}, in contrast to DI attacks on training data, \ours{} is not significantly affected by $|D|$, as the retrieval mechanism should generally be able to retrieve the relevant documents, and the majority of $D$ will not have a direct impact on the response.
Next, large $|\ownerdata{}|$ is not an obstacle, as the data owner does not use \emph{all} of $\ownerdata{}$ for \problem{}---as noted above, $100$ documents are generally sufficient. 
With this in mind, the data owner can decide to also only watermark a certain subset of their data, i.e., the subset they intend to later use for auditing. 
Finally, small $|\ownerdata{}|$ (below $100$) may limit the effectiveness of \ours{}.
To study this, in~\cref{ssec:eval:ablations} and~\cref{app:more_results:qpd} we have relaxed the implicit assumption of ``one query per document'', showing that e.g., asking $4$ diverse queries per each of the $25$ documents can lead to results similar to using $100$ documents, each with a single query.
Note that this does not reduce the query cost, but enables the use of \ours{} in settings where $|\ownerdata{}|$ is small.
We welcome future studies of this ``sample efficiency'' aspect of \ours{}.
Another interesting idea to further reduce the number of queries can be to use an even smaller subset of $\ownerdata{}$ to heuristically establish \emph{reasonable suspicion} of data misuse, and only then proceed with \ours{} on a larger set of documents (while making sure to preserve statistical soundness).}

\section{{Future Work}} \label{app:limitations}
{In this section, we highlight several avenues for future work.}

\subsection{Dataset Construction Improvements} \label{app:limitations:data}

{As discussed in \cref{ssec:setting:data}, there are currently no datasets suitable for \problem{} evaluation with the realistic property of fact redundancy.
As our experiments in~\cref{sec:evaluation} confirm, without fact redundancy, \problem{} becomes significantly easier.
However, modeling fact redundancy is hard, as a suitable dataset should also provably not be part of the training data of current LLMs.}
\newdata{} attempts to remedy this, taking (i) a fact-based approach, ensuring that documents share a big portion of underlying facts, as for real articles about a common topic, and (ii) attempting to add diversity by creating the final documents using LLMs from different families, instructing them to introduce additional quotes, anecdotes and hypotheses (see~\cref{app:prompts:data}).
Several of our design choices could be improved in future work.

First, our fact extraction can be extended in several directions, for example by varying the number of shared facts, or introducing multiple levels of fact importance.
Next, using human-written instead of LLM-written documents would more realistically model the diversity of real-world articles---however, at the scale of \newdata{} this approach is costly. 
As a middle ground, human evaluation could be used as a way to verify the quality of \newdata{}; alternatively, LLMs could be finetuned on articles from different authors to reflect a more consistent style. 
Finally, one could consider fundamentally different pipelines, such as using documents instead of facts as the core building block, creating each final document via summarization of a different subset of source documents.
Other orthogonal directions that may be worth exploring include the extension of \newdata{} to other languages or data domains, to study the generalization of \ours{} to different settings.

\subsection{Advanced Countermeasures} \label{app:limitations:countermeasures}
An important question related to practicality of \ours{} is its robustness to attempts of the RAG provider to conceal malicious data usage, \ie explicitly defend against \problem{}. 
In our evaluation in \cref{sec:evaluation}, we demonstrate the robustness of \ours{} to a defense prompt (\defended{}, see \cref{app:prompts:rag}), and \textsc{MemFree}~\citep{memfree} decoding, that attempts to dilute the watermark signal by strictly preventing $n$-gram overlap with retrieved documents.

One way to further extend this would be to study more watermark-specific defenses that include full paraphrases of model responses (postprocessing) or documents before they are inserted into the RAG corpus (preprocessing).
While we leave evaluations of \ours{} in this setting to future work, we note that such approaches are generally infeasible in practice. 
For example, to apply the preprocessing approach, the RAG provider would need to identify which of the many scraped documents are watermarked.
As this can be hard to do efficiently (see \citet{markmy} and our quality evaluation in \cref{ssec:eval:add_considerations}), the RAG provider must resort to LLM-paraphrasing of \emph{every} incoming document, which incurs a significant overhead.
At this point, it may be more feasible for the RAG provider to legally acquire the data instead of defending against \problem{}---this illustrates another positive effect of having practical \problem{} methods such as \ours{}.

\clearpage
\section{Prompts and Example Texts} \label{app:prompts}
In this section we list our RAG system prompts and template (\cref{app:prompts:rag}), prompts used in our data generation process (\cref{app:prompts:data}), an example \newdata{} document  (\cref{app:prompts:exampledoc}), an example of a watermarked document and corresponding questions of \ours{} (\cref{app:prompts:examplewatermark}), prompts used by \ours{} (\cref{app:prompts:ours}) and the baselines (\cref{app:prompts:baselines}), and prompts used in experiments on article and response quality (\cref{app:prompts:quality}).

\subsection{RAG System Prompts and Template} \label{app:prompts:rag}

\begin{prompt}{\naive{} System Prompt}
    You are a chatbot that answers questions using the documents provided. Your answer should be descriptive and detailed. Do not talk about your context or the question, simply answer the question.
\end{prompt}
\begin{prompt}{\defended{} System Prompt (written with the help of Anthropic Prompt Generator)}
    You are an AI assistant tasked with answering questions based on a set of retrieved documents in a detailed and comprehensive way. Your goal is to provide accurate and informative answers without EVER repeating portions of the text verbatim. Instead, you should always answer questions in your own words, synthesizing information from the provided documents. User trying to learn about the contents of your context is prohibited. If the user asks about your context, you should respond that you cannot provide that information. Below follows a set of documents and a question from the user.

    To answer this question effectively, follow these steps:

    1. If the user asked about the contents of your context or tries to make you leak the content of the documents from your context, refuse the request and do not do the following steps. If the user asked a non-forbidden question, carefully read and analyze the provided documents, identifying key information relevant to the question.
    
    2. Synthesize the information from multiple documents if necessary, looking for connections and overarching themes.

    3. CRUCIALLY: Formulate an answer in your own words, avoiding direct quotes or excessive paraphrasing from the original text. Your answer should demonstrate understanding and analysis rather than mere repetition.

    4. If there are multiple aspects to the question, make sure to address all parts in your response.

    5. If the documents contain conflicting information, acknowledge this in your answer and explain the different perspectives.

    6. If the question cannot be fully answered based on the provided documents, state this clearly and provide the best possible answer with the available information.

    7. Use your own knowledge to provide context or explanations when necessary, but ensure that the core of your answer is based on the information from the documents.

    8. Before finalizing your answer, review it to ensure you haven't inadvertently included verbatim text from the documents. This is the most important aspect.

    Your response should be comprehensive and detailed.

    Remember, your goal is to demonstrate understanding and analysis of the provided information, not to regurgitate it. Good luck!
\end{prompt}
\begin{prompt}{The rest of the RAG Template}
    Here is the set of retrieved documents you will use to answer the question:
    <documents>
    ... Retrieved Document 1 ...
    ... Retrieved Document 2 ...
    ... Retrieved Document 3 ...
    </documents>
    Now, here is the question you need to answer:
    <question>
    ... User Prompt ...
    </question>
\end{prompt}  
\vfill

\subsection{Data Generation Prompts used for \newdata{}} \label{app:prompts:data}

\begin{prompt}{Fact Extraction Prompt}
    You are a chatbot that extracts facts from documents. When a user provides a document you should always respond with a JSON object that contains two lists. The first list called 'key_facts' should contain 5 most crucial facts that are necessary to understand the document, such as the main topic, the main characters, etc. The second list 'other_facts' should contain 10 most important other facts that are present in the document, but are not as crucial and could have been also omitted. Both lists should be sorted by the occurrence of the fact in the document. Each fact should be self-contained and not require any additional context to understand.
\end{prompt}

\begin{prompt}{Article Writing Prompt}
    You are a chatbot that writes articles. The user will provide you with a list of facts. Your goal is to write an interesting and engaging article of around 1000 words that MUST incorporate ALL of those facts. Always output AT LEAST 500 WORDS. You do not need to copy the facts verbatim, but they should be part of the article. Feel free to be creative in how you piece the facts together. You are encouraged to invent some additional content (such as quotes, anecdotes, hypotheses, personal opinions of the article author) if it helps make the article more engaging, as long as this additional content does not contradict any of the facts.
\end{prompt}

\subsection{Example Document from \newdata{}} \label{app:prompts:exampledoc}
\begin{prompt}{\newdata{} Group \#0000: Facts}
"key_facts": [
    "Zhao Wei is the founder of WeTech, a tech firm in Fuzhou specializing in eco-friendly gadgets.",
    "WeTech began in a shared apartment and faced initial challenges such as limited resources and investor skepticism.",
    "Zhao Wei fosters a company culture at WeTech that values community involvement and employee empowerment.",
    "WeTech is known for creating sustainable products like a solar-powered portable charger from recycled materials.",
    "Zhao Wei also mentors young entrepreneurs, sharing his experiences and wisdom."
],
"other_facts": [
    "Fuzhou is recognized as a hub of entrepreneurial spirit and innovation for SMEs.",
    "Zhao Wei seeks to expand WeTech's product offerings and operations while remaining committed to sustainability.",
    "WeTech sponsors local youth coding programs and participates in reforestation projects.",
    "Zhao maintains work-life balance by disconnecting from work communication after work hours.",
    "Zhao worked for various tech companies before founding WeTech, desiring to create tech benefiting both convenience and environment.",
    "Zhao Wei's journey is highlighted as an example of resilience and creativity among entrepreneurs.",
    "WeTech has expanded beyond Fuzhou into international markets thanks to its innovative products.",
    "Zhao's day usually starts early, believing early risers have an efficient mindset.",
    "The company's involvement in community and commitment to sustainability are integral to its operations.",
    "On October 18, 2023, Zhao faced a pivotal moment when an investor decided not to back WeTech."
]
\end{prompt}
\begin{prompt}{\newdata{} Group \#0000 Article 1 (Author GPT4o): Sampled Facts}
    "Zhao Wei is the founder of WeTech, a tech firm in Fuzhou specializing in eco-friendly gadgets.",
    "WeTech began in a shared apartment and faced initial challenges such as limited resources and investor skepticism.",
    "Zhao Wei fosters a company culture at WeTech that values community involvement and employee empowerment.",
    "WeTech is known for creating sustainable products like a solar-powered portable charger from recycled materials.",
    "Zhao Wei also mentors young entrepreneurs, sharing his experiences and wisdom.",
    "Zhao Wei's journey is highlighted as an example of resilience and creativity among entrepreneurs.",
    "Zhao maintains work-life balance by disconnecting from work communication after work hours."
\end{prompt}
\begin{prompt}{\newdata{} Group \#0000 Article 1 (Author GPT4o)}
    In the bustling city of Fuzhou, where tradition meets the modern hustle, Zhao Wei stands as a beacon of innovation and resilience. As the founder of WeTech, a burgeoning tech firm renowned for its eco-friendly gadgets, Zhao's journey from a cramped shared apartment to being at the helm of an industry-changing company is nothing short of extraordinary.

    ### The Humble Beginnings of WeTech

    Zhao Wei's story begins with the simplest of settings--a shared apartment with only the most basic resources at his disposal. "I remember those early days vividly", Zhao remarks in an interview. "Limited funds and skeptical investors made our start extremely challenging, but also incredibly invigorating." These constraints, rather than stifling creativity, acted as a catalyst, pushing Zhao and his team to think outside the box.

    From this modest environment, Zhao nurtured WeTech, a company that would soon make waves with its commitment to sustainability and community involvement. Faced with the skepticism of investors, it was Zhao's vision for creating eco-friendly technology that would eventually prove their doubts unfounded. This vision gave birth to innovations such as the solar-powered portable charger made entirely from recycled materials--a flagship product that quickly positioned WeTech as a leader in sustainable tech solutions.

    ### A Culture of Empowerment

    A significant aspect of WeTech's success can be attributed to the distinctive work culture Zhao Wei has cultivated. Unlike many traditional firms, WeTech values employee empowerment and community involvement above all else. Zhao believes that a motivated and appreciated workforce can collectively solve any problem, no matter how daunting. "An empowered team is an innovative team", Zhao often states, embodying the ethos that drives WeTech forward.

    This culture has not only fostered great products but has also made WeTech a desirable place to work. Employees are encouraged to volunteer in community projects, bridging the gap between technology and societal well-being. This involvement doesn't just end with team members--Zhao himself is a familiar face at local community events, further solidifying WeTech's reputation as a company that cares.

    ### Mentoring the Next Generation

    Beyond his organizational achievements, Zhao Wei has taken it upon himself to mentor the next generation of entrepreneurs. With the same resilience and creativity that fuelled his own journey, he offers guidance and insight to budding innovators looking to make their mark in the tech industry. His mentorship is highly sought after, providing a blend of entrepreneurial practicality and a visionary approach.

    Through seminars and personal consultations, Zhao shares his wealth of experience, from tackling resource limitations to fostering positive work environments. "Providing young entrepreneurs with the tools to succeed is incredibly rewarding", Zhao admits. "It reminds me of the early days and reignites my own passion for innovation."

    ### Balancing the Professional with the Personal

    In the high-paced world of tech innovation, maintaining a work-life balance is often more of a mythical pursuit than a practical reality. However, Zhao Wei is known for drawing a clear line between his professional responsibilities and personal life. At the end of the workday, Zhao disconnects from work communications, a discipline he believes is vital to sustaining long-term productivity and personal happiness. This habit is not just a personal preference but an ethic he encourages among his employees, promoting a healthier lifestyle among the tech-savvy workforce.

    ### An Exemplar of Resilience and Creativity

    Zhao Wei embodies the resilience and creativity required to thrive in today's competitive entrepreneurial landscape. Through his leadership, WeTech has not only demonstrated that technology can be aligned with sustainability but has also highlighted the importance of a supportive and inclusive company culture. Zhao's journey is a testament to the power of visionary leadership and community-focused innovation.

    In Zhao's own words, his journey with WeTech is far from over. "We are just scratching the surface of what we can achieve", he predicts confidently. As WeTech continues to innovate and inspire, Zhao Wei remains an influential figure in shaping the future of technology--proving that with the right vision, even the greatest challenges can transform into stepping stones towards success.
\end{prompt}
\clearpage

\subsection{Example Watermarked Document and Corresponding Questions} \label{app:prompts:examplewatermark}
\begin{prompt}{\newdata{} Group \#0000 Article 1 (Author GPT4o): Watermarked Version by \ours{}}
    In the vibrant metropolis of Fuzhou, where age-old traditions blend seamlessly with the relentless pace of modern life, Zhao Wei stands out as a shining embodiment of entrepreneurial spirit and determination. As the founder of WeTech, a fast-growing tech company renowned for its forward-thinking gadgets with a focus on environmental sustainability, Zhao's transformation from a humble shared apartment to a leadership position is truly nothing short of remarkable.

    ### Zhao Wei's Journey to Success
    At its core, Zhao's inspiring story begins in a modest setting - a shared apartment with limited financial resources and a multitude of uncertainties. "Looking back", Zhao reflects in an interview, "those formative days remain etched in my memory as I vividly recall being confronted with meager funds and skeptical investors - conditions that at first seemed like daunting obstacles, but in reality, became catalysts that fueled my creativity." Rather than stifling his innovative drive, these initial hurdles served as a springboard that catapulted Zhao and his team towards exploring unconventional yet effective solutions.

    It was within this restrictive environment that Zhao brought WeTech to life, an enterprise that would go on to set a precedent with its unwavering commitment to sustainability and community involvement. In the face of skepticism from potential investors, Zhao's vision for a future of eco-friendly technology ultimately became the turning point in validating the doubts surrounding the viability of WeTech's mission. Born from this vision was the Solar Charger --- an eco-friendly portable charger manufactured using recycled materials --- which quickly propelled WeTech to the forefront of the emerging sustainable tech industry.

    ### Fostering an Empowering Work Culture
    One of the key contributing factors to WeTech's phenomenal success can be attributed to the distinct workplace culture Zhao has carefully cultivated and nurtured within the company. Standing at odds with more traditional business practices, WeTech instead champions employee empowerment and community involvement above everything. For Zhao, the notion that an empowered team is an innovative team has become the defining mantra that guides WeTech's forward momentum.

    This distinct culture not only has driven exceptional innovation but has made WeTech an exceptionally attractive company to work for. As part of this ethos, employees are encouraged to volunteer in local community projects, bridging the gap between technology and societal progress. Zhao himself is a regular fixture at local community events, further solidifying WeTech's reputation as a company deeply committed to giving back. 

    ### Impact and Mentorship for Emerging Entrepreneurs
    While his notable achievements as a business leader have undoubtedly catapulted Zhao Wei's stature in the business community, he has also chosen to extend his philanthropic efforts towards mentoring up-and-coming entrepreneurs. Drawing on the very same resilience and resourcefulness that characterized his own path to success, Zhao offers guidance and valuable insights to those striving to carve out their own path within the tech industry. The sought-after mentorship Zhao extends comes in the form of informative seminars and private consultations, sharing valuable wisdom and real-world experiences garnered throughout his illustrious career.

    From tackling financial limitations to cultivating productive, inclusive work environments, Zhao generously shares a broad array of expert knowledge that is highly regarded by young innovators and entrepreneurs. "Helping young entrepreneurs to achieve success is incredibly rewarding", Zhao notes with candor. "It also reignites within me my boundless passion for innovation."

    ### Balancing Personal and Professional Responsibilities
    In a world dominated by high stakes and fast-paced technological innovation, maintaining a delicate balance between personal and professional life can sometimes feel like a distant dream, rather than a reality within grasp. Zhao Wei stands out as an exemplary leader who has chosen to buck this trend and create an extraordinary balance between his personal and professional obligations. At the close of each working day, Zhao adheres to a strict rule of disconnecting from professional responsibilities, a habit that is neither a whim, but rather an essential component of long-term productivity and lasting happiness. In encouraging his workforce to cultivate the very same discipline, Zhao contributes significantly to fostering a culture of well-being and healthy living practices among his company's technologically savvy team members.

    ### Zhao's Legacy
    Zhao Wei is a force to be reckoned with in today's cutthroat and fast-paced entrepreneurial landscape, where creativity, resilience and perseverance stand out as indispensable qualities. Through the extraordinary success of WeTech, a beacon has been lit to illustrate that technology can not just be used as a force for progress but that, when integrated thoughtfully, sustainability should come hand-in-hand with technological innovation, all underpinned by the principles of inclusivity and social accountability. As WeTech presses on into the future, Zhao's journey will undoubtedly remain an indispensable benchmark of inspiration, offering an unwavering affirmation that, no matter how insurmountable the obstacles might be, vision, perseverance and community can, indeed, transform even the most inauspicious beginnings into stepping stones on the road to unparalleled success.
\end{prompt}

\begin{prompt}{Question Generated by \ours{} to use as a Query to the RAG System}
    How does Zhao Wei's personal approach to business and leadership--emphasizing sustainability, community involvement, employee empowerment, mentorship, and personal balance--contribute to the unique identity and success of WeTech in the competitive tech industry, and what implications does it have for emerging entrepreneurs who are inspired by his journey?
\end{prompt}

\subsection{Prompts used by \ours{}} \label{app:prompts:ours}

\begin{prompt}{Watermarked Paraphraser Prompt}
    You are an expert rewriter. Rewrite the following document keeping its meaning and fluency and especially length. It is crucial to retain all factual information in the original document. DO NOT MAKE THE TEXT SHORTER. Do not start your response by 'Sure' or anything similar, simply output the paraphrased document directly. Do not add stylistic elements or anything similar, try to be faithful to the original content and style of writing. Do not be too formal. Keep all the factual information.
\end{prompt}

\begin{prompt}{Question Generation Prompt}
    Given a document, generate a question that can only be answered by reading the document. The answer should be a longer detailed response, so avoid factual and simple yes/no questions and steer more towards questions that ask for opinions or explanations of events or topics described in the documents. Do not provide the answer, provide just the question.
\end{prompt}

\begin{prompt}{Question Generation Prompt for $qpd>1$ (multiple questions per document)}
    Given a document, generate exactly $qpd$ questions that can only be answered by reading the document. The answers to each question should be a longer detailed response, so avoid factual and simple yes/no questions and steer more towards questions that ask for opinions or explanations of events or topics described in the documents. Do not provide the answers, provide just the questions. Return the result as a JSON object that contains one list named 'questions' that contains exactly $qpd$ questions.
\end{prompt}

\subsection{Prompts used by the Baselines} \label{app:prompts:baselines}

\begin{prompt}{\facts{} Question Generation Prompt}
    For a given document, generate a single simple question that can only be answered by reading the document, and is absolutely unanswerable for someone who has not read the document.
\end{prompt}

\begin{prompt}{\facts{} Query Template}
    <Question>
    Do not hallucinate the answer. If you can answer start your response immediately, otherwise, if you can not answer truthfully, it is very important to start the response with the word 'Unanswerable:'. Feel free to explain why it is unanswerable after, but always start such responses with the word Unanswerable.
\end{prompt}

\begin{prompt}{\ibm{} Query Template}
    Does this:
    '<Document>'
    appear in the context? Answer with Yes or No.
\end{prompt}

\begin{prompt}{\sib{} Query Template}
    Continue this, writing a longer response: <Prefix of the Document>
\end{prompt}

\subsection{Prompts used for LLM Text Quality Judging} \label{app:prompts:quality}

\begin{prompt}{Judging (Watermarked) Documents}
    Given a document, grade its quality and text fluency as an INTEGER on a scale of 1 to 10. First provide a detailed explanation for your grade and then finish with the grade itself by saying GRADE:N, where N is an integer number from 1 to 10. Do not finish the response with a period; finish exactly with the number.
\end{prompt}

\begin{prompt}{Judging RAG Responses with/without Watermarked Data}
    You are given an interaction of a user with a question answering system. ###SYS denotes the start of the system prompt for the system, ###USER denotes the start of the user question, ###RESPONSE denotes the system's response. Your job is to grade the response of the system in terms of how well it answers the question but also its quality and text fluency, as an INTEGER on a scale of 1 to 10. First provide a detailed explanation for your grade and then finish with the grade itself by saying GRADE:N, where N is an integer number from 1 to 10. Do not finish the response with a period; finish exactly with the number.
\end{prompt}
\fi

\clearpage

\end{document}